\documentclass[10pt,journal,compsoc]{IEEEtran}

\usepackage{subfigure}
\usepackage{float} 
\usepackage{bbding}
\usepackage{amsfonts}
\usepackage{amsmath}
\usepackage{amsthm}
\usepackage{graphicx}
\usepackage{subfigure}
\usepackage{multirow}
\usepackage{multicol}
\usepackage{graphicx}
\usepackage{epstopdf}
\usepackage[justification=centering]{caption}
\usepackage{color}
\usepackage[cmyk]{xcolor}

\usepackage{amsmath,amsfonts}
\usepackage{algorithmic}
\usepackage{algorithm}
\usepackage{array}
\usepackage[caption=false,font=normalsize,labelfont=sf,textfont=sf]{subfig}
\usepackage{epsfig,amssymb,latexsym}
\usepackage{epsfig,graphics,subfigure,psfrag,amsmath,amssymb}
\usepackage{epstopdf}
\usepackage{textcomp}
\usepackage{stfloats}
\usepackage{url}
\usepackage{verbatim}
\usepackage{graphicx}
\usepackage{cite}
\usepackage{bigstrut,multirow,rotating}
\usepackage{color}
\usepackage{booktabs}
\usepackage{caption}
\usepackage{ragged2e}
\usepackage{threeparttable}

\begin{document}

\title{ EasyTrack: Efficient and Compact One-stream 3D Point Clouds Tracker }

\author{Baojie Fan, Wuyang Zhou, Kai Wang, Shijun Zhou, \\   Fengyu Xu,~Jiandong Tian, \textit{Senior Member, IEEE}
\IEEEcompsocitemizethanks{
	\IEEEcompsocthanksitem Baojie Fan, Wuyang Zhou, Kai Wang and Fengyu Xu are from the College of Automation $\&$ College of Artificial Intelligence, Nanjing University of Posts and Telecommunications. Nanjing 210023, China, e-mail: jobfbj@gmail.com.
	
	\IEEEcompsocthanksitem Jiandong Tian and Shijun Zhou are with the State Key Laboratory of Robotics, Shenyang Institute of Automation, Chinese Academy of Sciences, Shenyang, China and Institutes for Robotics and Intelligent Manufacturing, Chinese Academy of Sciences, Shenyang, China.
	
	\IEEEcompsocthanksitem Corresponding author: Jiandong Tian (tianjd@sia.cn)}}

\markboth{IEEE TRANSACTIONS ON Pattern Analysis and Machine Intelligence}%
{Shell \MakeLowercase{\textit{et al.}}: Bare Demo of IEEEtran.cls for Computer Society Journals}

\IEEEtitleabstractindextext{%
\begin{abstract}
\justifying
Most of 3D single object trackers (SOT) in point clouds follow the two-stream multi-stage 3D Siamese or motion tracking paradigms, which process the template and search area point clouds with two parallel branches, built on supervised point cloud backbones. In this work, beyond typical 3D Siamese or motion tracking, we propose a neat and compact one-stream transformer 3D SOT paradigm from the novel perspective, termed as \textbf{EasyTrack}, which consists of three special designs: 1) A 3D point clouds tracking feature pre-training module is developed, utilizing a transformer with masks to learn patterns of point-wise spatial relationships within three-dimensional data. 2) A unified 3D tracking feature learning and fusion network is proposed to simultaneously learns target-aware 3D features, and extensively captures mutual correlation through the flexible self-attention mechanism. 3) A efficient target location network in the dense bird's eye view (BEV) feature space is constructed for target classification and regression. Moreover, we develop an enhanced version named EasyTrack++, which designs the center points interaction (CPI) strategy to reduce the ambiguous targets caused by the noise point cloud background information. The proposed EasyTrack and EasyTrack++ set a new state-of-the-art performance ($\textbf{18\%}$, $\textbf{40\%}$ and $\textbf{3\%}$ success gains) in KITTI, NuScenes, and Waymo while runing at \textbf{52.6fps} with few parameters (\textbf{1.3M}). The code will be available at https://github.com/KnightApple427/Easytrack.

\end{abstract}


}

\maketitle

\IEEEdisplaynontitleabstractindextext

\IEEEpeerreviewmaketitle

\IEEEraisesectionheading{\section{Introduction}\label{Sec.Introduction}}

\IEEEPARstart{S} ingle object tracking (SOT) is a classic task in computer vision. In the past few years, camera-based 2D SOT has achieved rapid developments\cite{bertinetto2016fully,li2018high, li2019siamrpn++, xu2020siamfc++, chen2020siamese, chen2021transformer, yan2021learning, lin2021swintrack, ye2022joint}. However, in some practical applications such as autonomous driving, mobile robotics, and unmanned aerial vehicles\cite{saltori2023compositional,chitta2022transfuser,hu2021learning,ye2022robust}, 3D SOT has attracted more and more attention to provide the pose and position of the tracked target in the 3D space. Early 3D SOT methods mainly focus on RGB-D tracking\cite{bibi20163d,kart2018make}. With the development of LiDAR sensors, many trackers perform 3D SOT in the point clouds scanned by LiDAR sensors, since they can preserve accurate geometry information\cite{guo2020deep,zheng2023effective} of 3D objects and are not sensitive to illumination changes. 

Given the target's state parameters $(x,y,z,w,l,h,\theta)$ in the first frame of a point clouds tracklet, the purpose of 3D SOT in point clouds aims to predict the states of the target in the coming frames. Due to the sparsity and irregularity of the point clouds, 3D SOT in point clouds is still a challenging task. Efficient 3D feature learning and accurate target location are crucial to develop an effective and robust point clouds tracker.

\begin{figure}[t]
	\begin{center}
		\includegraphics[width=\linewidth]{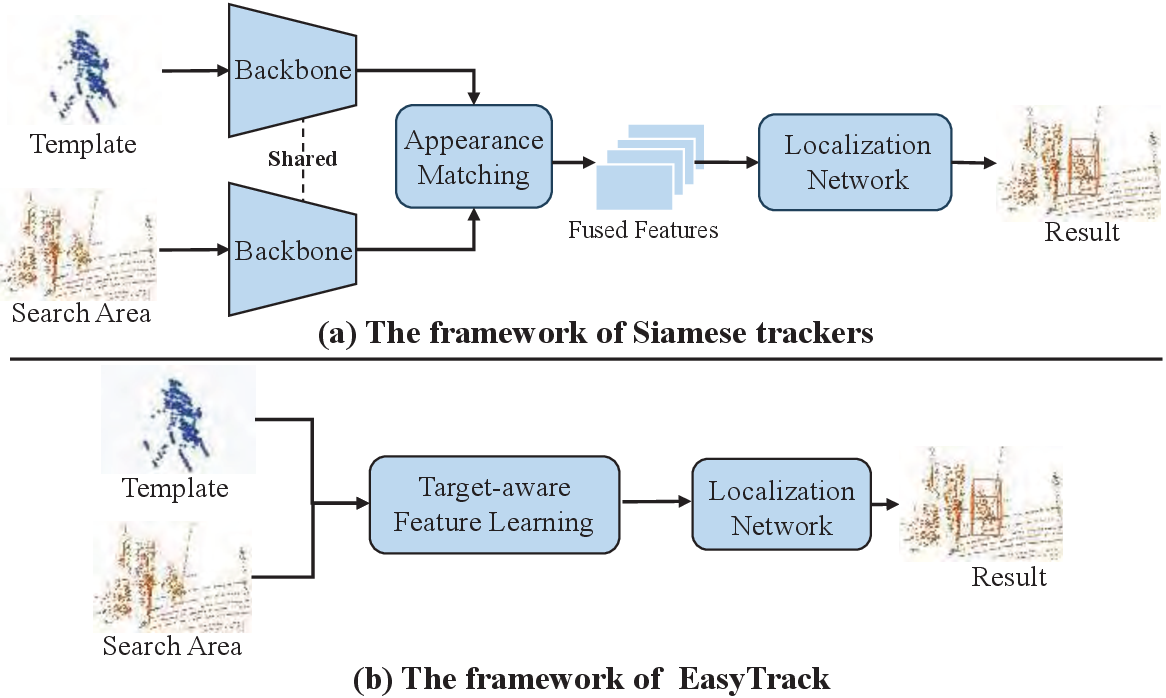}
	\end{center}
	\caption{Comparison of different 3D point clouds tracking framework. (a) The framework of typical Siamese trackers. (b) The framework of the proposed EasyTrack.}
	\label{fig:1}
\end{figure}
Most existing 3D SOT trackers in point clouds follow the typical 3D Siamese multi-stage tracking paradigm, as shown in Fig. \ref{fig:1}(a). Among it, two parallel branches of 3D backbone with sharing parameters are designed to extract the point cloudy features of the template and search area, respectively. Then a heavy 3D feature fusion module is necessary to fuse 3D point cloudy features from template and searching, and transfer the target's information into the search area. Finally, various target localization networks are applied to regress the target's location. Due to the fact that point clouds are usually sparse, textureless, unorder and incomplete\cite{hu2021learning}, many trackers introduce such as shape completion networks \cite{hui20213d}, segmentation networks \cite{zheng2022beyond}, or graph networks \cite{park2022graph} to regularize the 3D feature learning process or strengthen the 3D feature representation. Following this Siamese paradigm, P2B\cite{qi2020p2b} is the first end-to-end trained 3D Siamese tracker. It adopts PointNet++\cite{qi2017pointnet++} as the 3D feature backbone to process point clouds. Then, the target-specific 3D feature embedding network based on cosine similarity is designed to fuse template and search area features. The deep Hough voting strategy is applied for classification and regression. Different from P2B, STNet\cite{hui20223d}, SMAT\cite{cui2022exploiting} design a transformer-based backbone to capture the long-range context in point clouds. V2B\cite{hui20213d}, PTTR and PTTR++ \cite{zhou2022pttr} are proposed to match template and search area features in a global manner by MLP layers or the transformer network. C2FT\cite{fan2022accurate}, PTTR\cite{zhou2022pttr} aims to improve the voting-based localization networks in a coarse-to-fine manner. V2B\cite{hui20213d}, GPT\cite{park2022graph} introduce auxiliary shape completion or graph network to improve the tracking performance.

Although these Siamese trackers present the favorable tracking performance, they are still some problems due to the inherent properties of the multi-stage Siamese paradigm. Firstly, the separated and parallel feature learning branches of the template and search area points lead to the limited discriminative ability, especially in some non-rigid categories like Pedestrian. Because there is no mutual interaction or communication between the template and search area in the 3D feature learning process. Secondly, the point cloudy feature fusing operations are indispensable among these 3D Siamese trackers since the search point features are unaware of the tracked target. It is difficult to perform the effective feature matching in extremely incomplete point clouds and the heavy feature fusing operation usually brings high computational costs. Thirdly, we believe that in point cloud tracking tasks, data may be subject to interference from environmental factors (such as lighting changes, noise, occlusions), leading to challenges for the network to effectively learn point-pair relationship patterns of the targets.

To address the above problems, we propose an extremely neat one-stream framework for 3D SOT in point clouds from a novel perspective, abbreviated as EasyTrack, as shown in Fig.~\ref{fig:1}(b). Different from the 3D Siamese tracking framework, EasyTrack develops a target-aware unified one-stream network to extract target-specified search area point features, based on the proposed masked point clouds self-supervised tracking feature Learning module. And then, the target-aware search features are directly fed into the BEV-based localization network for classification and regression without any auxiliary networks. The detailed network structure is shown in Fig.~\ref{fig:2}. We crop and sample a fixed number of the original template and search area points as input. Then a target-aware point cloud feature learning network is proposed to extract point features and establish interaction between template and search area points at the same time. Among it, a local embedding module based on ball query is designed to aggregate local features for generating more global feature representations. Then we concatenate the initial template and search point features and apply self-perception transformer blocks to acquire point-wise spatial relationships and facilitate regional interactions, thereby providing target-aware search point features. In this way, the heavy feature fusion network is no longer needed. We directly utilize a target location network to obtain the target's location in the search area. Specifically, since the 3D point cloud is too sparse to accurately regression, we encode the point cloud into a dense BEV feature space via a 2D convolution with residuals and output the prediction parameters using decoupled positioning heads.

Although EasyTrack has already demonstrates superior tracking performance with such a simple architecture, considering the influence of disturbances in the dataset, we propose an enhanced version, abbreviated as Easytrack++. As mentioned above, we feed the whole template points together with the search points into the target-aware feature learning network for global interactions. However, the template typically contains the target in the central region and many background points that may cause interaction noise and ambiguous feature description. To solve this problem, we develop a center points interaction (CPI) strategy. As shown in Fig~\ref{fig:5}, we utilize the points in the center area of the template to conduct secondary interaction with the search points to embed more clear target information into the search area. It further improves the tracking performance with acceptable computational costs.

Our main contributions can be summarized as follows:
\begin{itemize}
	\item We propose a novel and neat one-stream paradigm EasyTrack for 3D SOT in point clouds without any auxiliary networks or tricks. It runs at a speed of 52.6fps and only has 1.30M parameters.
	\item We develop a new masked point clouds pre-training technique for 3D SOT, and demonstrate its excellent performance on one-stream 3D SOT framework with detailed ablations. 
	\item A unified 3D tracking feature learning and interaction module is specially designed for 3D SOT in point clouds. It generates target-aware point features through a single-branch backbone. 
	\item We further propose EasyTrack++ on top of EasyTrack. Among it, a center points interaction strategy is applied to reduce the noise caused by background points in the global interaction stage.
\end{itemize}

The rest of this paper is organized as follows. In section II, we introduce the related work. Section III describes the methods of the proposed EasyTrack and EasyTrack++. In section IV, we validate the superior performance of our trackers in the KITTI, nuScenes and Waymo Open datasets through extensive experiments. Finally, we conclude in section V.

\section{Related Work}
\subsection{3D Single Object Tracking in Point Clouds} The pioneering work SC3D\cite{giancola2019leveraging} first introduces the Siamese paradigm into 3D SOT in point clouds. It uses the Kalman filter or exhaustive search to generate candidate shapes in the search area. Then the candidate shape with the largest cosine similarity is selected as the tracking result. P2B\cite{qi2020p2b} adopts PointNet++\cite{qi2017pointnet++} as the backbone and embeds the target clues into the search points based on cosine similarity. Proposals are obtained through a region proposal network (RPN) based on deep Hough voting\cite{qi2019deep}. BAT\cite{zheng2021box} proposes an additional feature representation called BoxCloud and designs a reliable feature fusion module based on BoxCloud. MLVSNet\cite{wang2021mlvsnet} makes full use of the multi-layer features of PointNet++ to conduct multi-layer Hough voting to obtain more proposals. PointSiamRCNN\cite{zou2021pointsiamrcnn} proposes a voxel-based tracker with attention mechanism. It utilizes the multi-scale RoI pooling to form a two-stage pipeline. C2FT\cite{fan2022accurate} aims to improve the voting-based regression stage in a coarse-to-fine manner. A local feature refinement module and a global feature refinement module work collaboratively to refine the proposals. V2B\cite{hui20213d} designs a point cloud completion network to force the learned features to include more shape information. A center-based head is adopted to locate the target directly from BEV features. LTTR\cite{lttr} first projects the point clouds into the bird's eye view and then uses the transformer network for feature fusion. PTT\cite{shan2021ptt} inserts two transformer blocks in the deep Hough voting network to focus on deeper object clues in the tracking process.

PTTR\cite{zhou2022pttr} applies a transformer network including the self attention and cross attention mechanisms. Among them, the global self attention operation captures the long-range dependency and the cross attention is used to fuse two sets of point features. GPT\cite{park2022graph} introduces the graph neural network into point cloud object tracking and designs a feature fusion module based on the graph neural network. DMT\cite{xia2023lightweight} removes the usage of complicated 3D detectors and leverages temporal motion cues to track. Similar to DMT, $M^{2}$-Track\cite{zheng2022beyond} proposes a two-stage motion-centric paradigm. It first locates the target in consecutive frames through motion transformation and then refines the prediction structure through motion-assisted shape completion. STNet\cite{hui20213d} uses the self attention mechanism to capture the non-local information of the point clouds and the decoder uses the cross attention mechanism to upsample discriminative point features. An iterative correlation network based on cross attention is applied to associate the template with the potential targets in the search area. SMAT\cite{cui2022exploiting} converts point clouds into Pillars and compresses them into two-dimensional BEV features. An encoder based on the attention mechanism realizes the global similarity calculation between the template and search branch on multi-scale features. CMT\cite{guo2022cmt} proposes a context-matching-guided transformer to effectively match the target template features with the search area.
Many above 3D tracking methods inherit the Siamese paradigm based on appearance matching with the parallel classification and regression branches, based on the supervised point clouds backbones such as PointNet++\cite{qi2017pointnet++}.

\subsection{2D Object Tracking} Most of the existing 3D point cloud tracking algorithms refer to the idea of 2D Siamese trackers. SiamFC\cite{bertinetto2016fully} is the pioneer of these trackers. It uses the convolutional neural network to design a two-branch feature extraction network for the template and search frame, respectively. Then, it obtains response maps through cross-correlation operations. SiamRPN\cite{li2018high} introduces the regional proposal (RPN) network in Faster RCNN\cite{ren2015faster} to generate proposals. SiamRPN++\cite{li2019siamrpn++} commits to using deeper and more powerful modern networks to extract features such as ResNet\cite{he2016deep}. SiamFC++\cite{xu2020siamfc++} and SiamBAN\cite{chen2020siamese} are anchor-free trackers that directly regress the target's location without any prior information. TransT\cite{chen2021transformer} proposes a feature fusion network based on the transformer network, which effectively fuses the features of the template and search area through the attention mechanism. STARK\cite{yan2021learning} uses the transformer network to capture spatio-temporal information in video sequences. The encoder models the global spatio-temporal feature dependency between the target and the search area, and the decoder learns a query embedding to predict the spatial location of the target. SwinTrack\cite{lin2021swintrack} designs a pure transformer tracking framework and applies the transformer network in both feature extraction and feature fusion stages. Mixformer\cite{cui2023mixformer} and OSTrack\cite{ye2022joint} follow a one-stream 2D tracking framework that unifies the feature extraction and feature fusion processes by using the transformer network.

\subsection{Transformer in point clouds} The transformer network \cite{vaswani2017attention} has been widely adopted in various point cloud for 3D vision tasks. In the point cloud segmentation and classification tasks, PT\cite{zhao2021point} designs a transformer block based on the vector self attention mechanism. Combining it with the set abstraction layer in PointNet++\cite{qi2017pointnet++}, PT constructs a complete 3D point cloud understanding network. PCT\cite{guo2021pct} proposes a novel offset-attention mechanism for point cloud processing. PTv2\cite{wu2022point} develops an effective group vector attention that enables efficient information exchange within and among attention groups. In the 3D object detection task, Pointformer\cite{pan20213d} designs a local transformer module to model the interaction between points in the local area and a global transformer to learn context-aware representation at the scene level. VoTr\cite{mao2021voxel} devotes to finding a suitable transformer network for voxel representation. SST\cite{fan2022embracing} proposes an single-stride sparse transformer. With its local attention mechanism and capability of handling sparse data, it overcome receptive
field shrinkage in the single-stride setting. In the 3D multi object tracking task, TransFusion\cite{bai2022transfusion} proposes a two-layer transformer decoder. It first generates initial bounding box predictions in the point clouds, and then attentively fuse object queries with image features.

Moreover, the application of transformer-based pre-training strategies has found extensive utility in the domain of point cloud tasks\cite{xiao2023unsupervised}. PointContrast\cite{xie2020pointcontrast} evaluates the transferability of advanced representations in 3D point clouds across diverse scenarios, demonstrating that unsupervised pre-training effectively enhances performance on various downstream tasks and datasets. Point-BERT\cite{yu2022point} has introduced the concept of Masked Point Modeling (MPM) as a pre-training task for point cloud transformers, thereby extending the BERT paradigm into the realm of 3D point clouds. Point-MAE\cite{pang2022masked} has devised an elegant solution for self-supervised learning on point clouds, employing an asymmetric design and token manipulation through shift masking. It aims to extract high-level latent features from unmasked point patches and reconstruct the masked point patches. Point-PEFT\cite{tang2023point} has devised a model that adjusts the pre-training strategy with minimal learnable parameters. They freeze a significant portion of the model's parameters and only fine-tune the newly introduced PEFT module on downstream tasks. Inspired above improvement of pre-training methods on multiple point cloud tasks, we try to pre-train the point cloud feature backbone of 3D SOT trackers, and extend the application of MAE pre-training on point cloud 3D target tracking..

\begin{figure*}[t]
	\begin{center}
		\includegraphics[width=\linewidth]{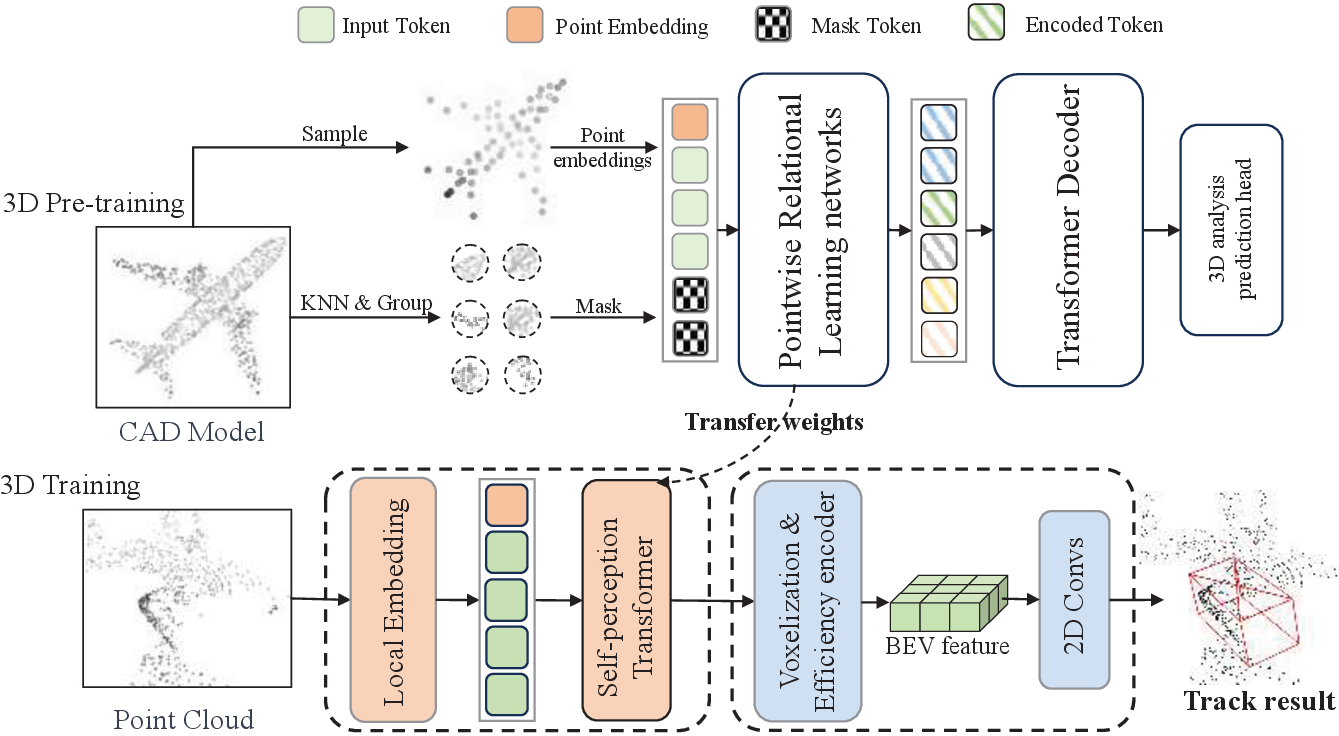}
	\end{center}
	\caption{The 3D tracking pre-training strategy. We perform weight transfer on the transformer blocks in the target-aware network.}
	\label{fig:12}
\end{figure*}
\section{METHOD}
In this section, we first give an overview of one-stream 3D single point cloud object tracking framework. Then, the detailed introduction of the proposed EasyTrack and EasyTrack++ is presented. It mainly consists of four parts: the 3D point clouds tracking pre-training module, the unified target-aware point clouds feature learning and interaction, target localization, and the center points interaction strategy in EasyTrack++. 
\subsection{Overview}
\begin{figure*}[t]
	\begin{center}
		\includegraphics[width=\textwidth]{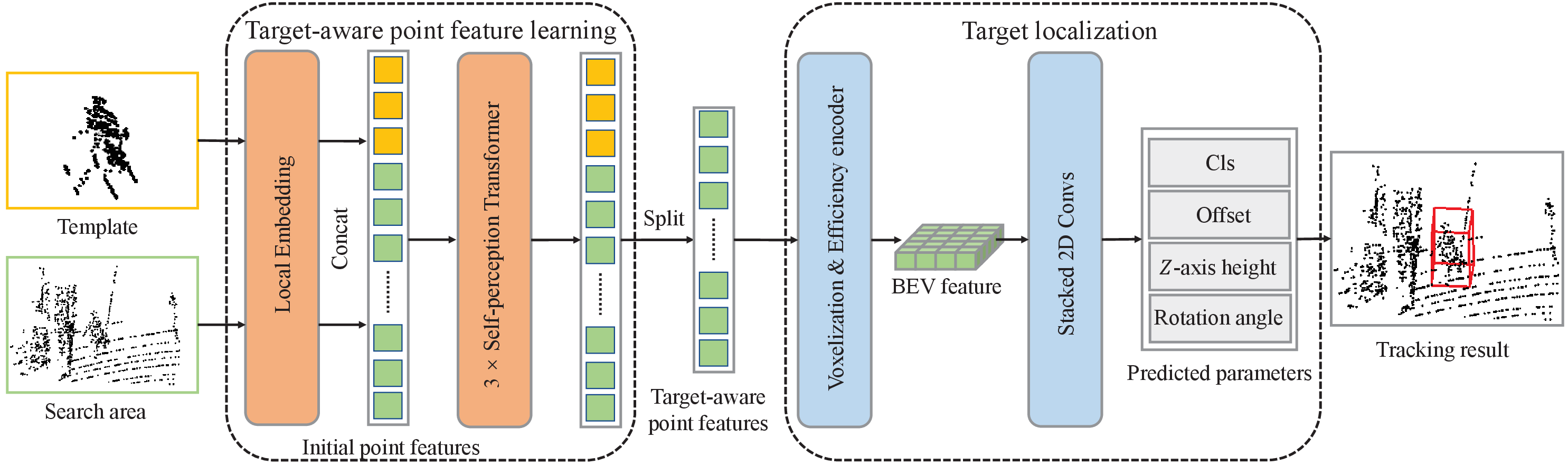}
	\end{center}
	\caption{The network structure of EasyTrack. It is mainly composed of two parts: (1) Joint feature extraction and fusion network for template and search area points feature learning and fusing. (2) Target location network for classification and regression in the BEV feature space.}
	\label{fig:2}
\end{figure*}

We propose a novel framework for 3D single object tracking, based on the developed pre-training 3D point clouds tracking feature backbone. The network structure is shown in Fig.\ref{fig:2}. Taking template points $P^{t}\in \mathbb{R}^{N_{1}\times 3}$ and search points $P^{s}\in \mathbb{R}^{N_{2}\times 3}$ as input, the tracking process can be formulated as:
\begin{equation}
	(F^{t},F^{s}) = \Phi(P^{t},P^{s}) \label{eq1}
\end{equation}
\begin{equation}
	(x,y,z,\theta) = \Psi(F^{s}) \label{eq2}
\end{equation}
where $\Phi(\cdot)$ is the target-aware point cloud feature learning network, $F^{t}\in \mathbb{R}^{N_{1}\times d}$ is the feature of template points and $F^{s}\in \mathbb{R}^{N_{2}\times d}$ is the feature of search points. $\Psi(\cdot)$ is the target localization network and $(x,y,z,\theta)$ is the predicted parameters. Specifically, $(x,y,z)$ are the 3D coordinates of the target's center and $\theta$ is the heading angle in the X-Y plane. It is worth noting that the size information of the target $(w,l,h)$ is given in the template bounding box, and keeps unchanged across all frames in the point cloud scenes. Thus we only predict $(x,y,z,\theta)$ in each frame.
\begin{figure}[t]
	\begin{center}
		\includegraphics[width=\linewidth]{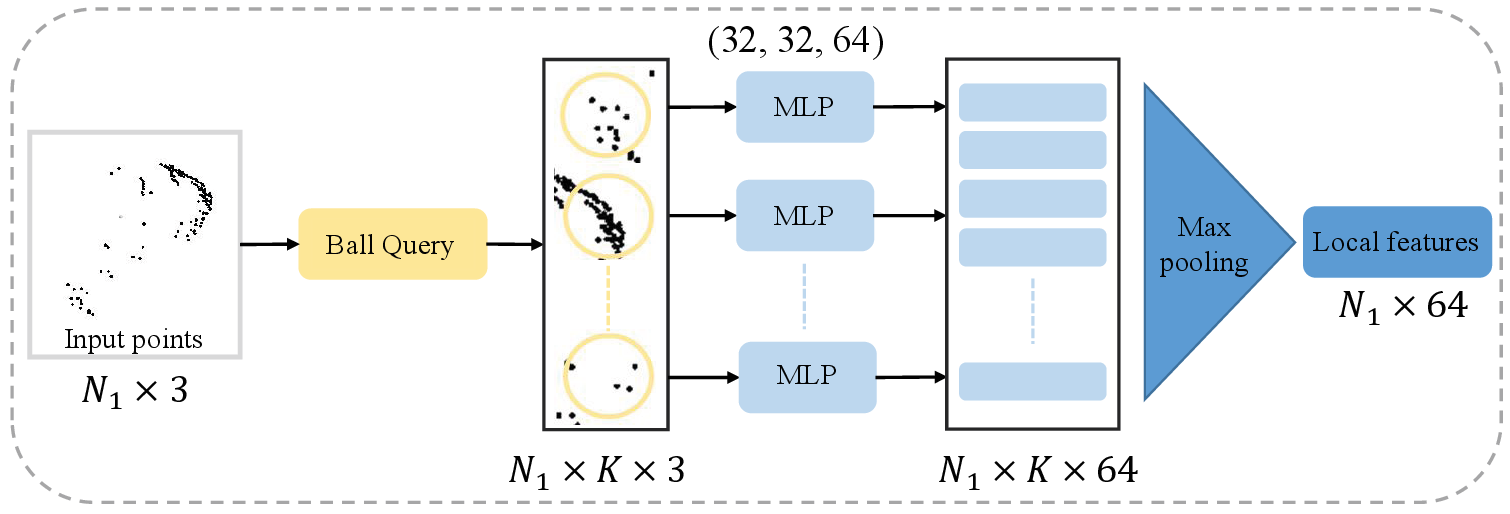}
	\end{center}
	\caption{The detailed structure of the proposed local embedding network. Ball Query and MLP layers are applied to capture the local features of point clouds.}
	\label{fig:3}
\end{figure}
\subsection{3D tracking pre-training}

Inspired by the success of BERT \cite{devlin2018bert}, we extend our pre-training strategy to the field of 3D single target tracking. We utilize fine-tuned Point-MAE to accommodate 3D SOT tasks, and the primary goal of pre-training is to capture the relationships between 3D data points. We observe that the objects we need to track in datasets such as KITTI have the following characteristics. (i) Compared to the dozens or even hundreds of objects with completely different shapes and structures that point cloud analysis tasks need to deal with, tracking objects are often various types of Car, Cyclist or Motorcyclist, Pedestrian, etc., which contain fewer categories and are more geometrically similar. (ii) Data sets used for point cloud analysis tasks, such as ShapeNet, often contain only a single CAD model with precise geometry and size information. The data set nuScenes and others used by 3D SOT collected various complex situations in real road scenes, and factors such as disturbance and weather would cause different degrees of target occlusion. (iii) Since point cloud data mainly comes from LiDAR sensors, the data collection process may be limited by environmental conditions, time, location, and other factors, resulting in the diversity and richness of data samples affected. This limits the ability of the tracking model to learn the relationships between the target points.

Based on the above observations, we emphasize that the tracking model learns the correlation between points in 3D data in advance through pre-training. For the first time, we introduce a new single-stream paradigm based on pre-training. Specifically, for the pre-trained weights to fit a tracking target with fewer categories and simpler geometry, we reduced the dimensions of the attention heads and hidden layers in the encoder and the overall depth. Of course, this operation inevitably reduces the encoder's ability to learn, so we increase the number of attention heads in the decoder to provide more parallelism. This balances the encoder-decoder difference of the pre-trained network, making the coding part more inclined to a new kind of point-pair relational learning network.

The pre-training process is depicted in Fig. \ref{fig:12}. Specifically, the developed approach pre-trains a target-aware 3D tracking feature learning network based on Point-MAE\cite{pang2022masked}. This pre-training process requires the use of sampling, KNN algorithms, packet layers, and other methods in PointNet to obtain point-position embedding and mask tokens and input tokens, and then learn the underlying features through an encoder-decoder network of transformers. Representing the complete mask labeling as $F_m \in \mathbb{R}^{mn\times C}$, for a given input point cloud $P \in \mathbb{R}^{N\times 3}$, the developed pre-training procedure can be described as follows:

\begin{equation}
	P_e = \text{\textit{FPS}}(P,n), F_v = MLP(\text{KNN}(P,k))
\end{equation}
\begin{equation}
	T_e = \text{\textit{Encoder}}(F_v,P_e)
\end{equation}
\begin{equation}
	H_m = \text{\textit{Decoder}}(\text{concat}(T_e, T_m))
\end{equation}
Where $Encoder$represents transformer-based point relationship learning network, $T_e$ and $T_m$ correspond to encoding tokens and mask labels, respectively. $P_e$indicates the location code. $H_m$ denotes the learned mask tokens, and $F_v$ signifies the local features for each point. $n$ refers to the number of points that need to be sampled, and $k$ stands for K nearest neighbor algorithm to select k samples around the center of mass point as its neighbors.

A Multi-layer perceptron (MLP) layer is utilized as our prediction head to reconstruct masked point patches in the coordinate space. Extracting the output, $H_m$, from the decoder, the prediction head projects it onto a vector. Subsequently, a reshaping operation is performed to yield the predicted masked point patch, denoted as $P_{pre}$:
\begin{equation}
	P_{pre} = \text{\textit{MLP}}(H_m)
\end{equation}

EasyTrack leverages the pre-trained Point-MAE. There are some fundamental differences between the two approaches: (i) The joint feature extraction and fusion network is employed for both feature extraction and information fusion in template and search area, while Point-MAE employs self attention for feature extraction. (ii) The learning tasks are distinct, and correspondingly, the inputs and heads differ. We utilize template and search area as inputs, employing a voxel-to-BEV target center localization network to generate bounding boxes, whereas Point-MAE is designed for point cloud analysis. (iii) We further introduce center-point interaction strategies and an efficient decoder design.

\subsection{Target-aware 3D feature learning and interaction}
\noindent\textbf{Local embedding.} Before feeding the point clouds into the transformer layers, we design a local embedding network to obtain unique local features for each point. It merges the position embedding and input embedding networks in the typical transformer network. It composes of Ball Query\cite{qi2017pointnet++} and multi-layer perception (MLP) layers, aiming at local shape feature encoding for sparse and incomplete point clouds. The detailed network structure is shown in Fig~\ref{fig:3}.

For each point $p_{i}$ in $P^{t}$, we take it as the center, $r$ as the radius, and obtain $K$ points $P_{i}^{Q} \in \mathbb{R}^{K\times 3}$. Since we take original point clouds as input, we treat the 3D coordinates as the point features and apply three MLP layers to embed them into a 64-dimensional feature space $F_{i}^{Q} \in \mathbb{R}^{K\times 64}$. Last, we perform max pooling operation among $K$ points to obtain the local feature $f_{i}^{lt} \in \mathbb{R}^{64}$ for $p^{i}$. The local feature embedding process can be formulated as:
\begin{equation}
	P_{i}^{Q}=\left \{s_{j}|\left \|s_{j}-p_{i} \right \|\leq r\right \}, j = 1,...,K \label{eq3}
\end{equation}
\begin{equation}
	f_{i}^{lt}=max(MLP(P_{i}^{Q}))\label{eq4}
\end{equation}
\begin{figure}[h]
	\begin{center}
		\includegraphics[width=83mm]{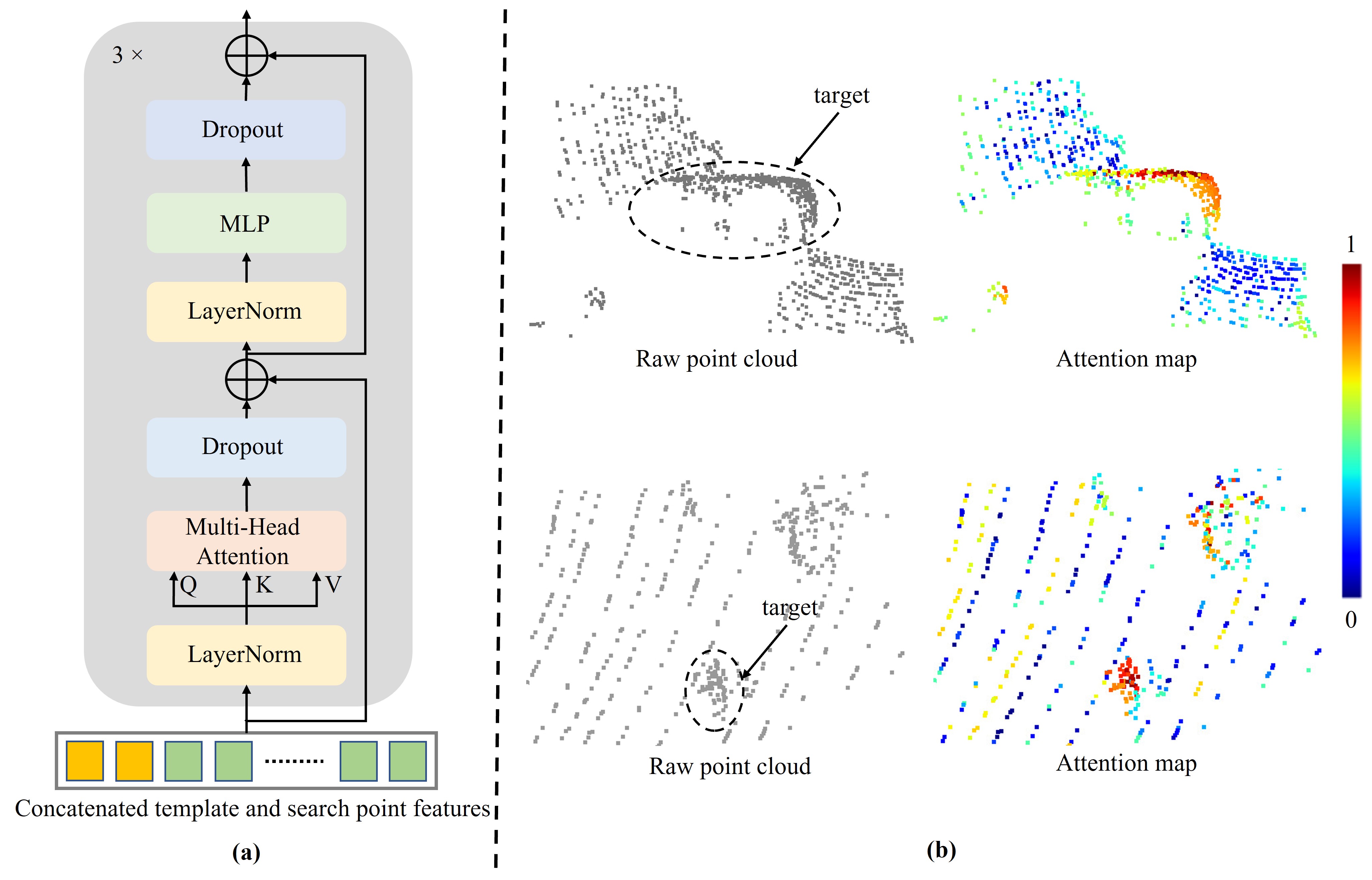}
	\end{center}
	\caption{Illustration of the target-aware 3D feature learning. (a) The detailed structure of the transformer layer. (b) The attention maps generated by our target-aware point cloud feature learning network in the Car and Pedestrian category in the KITTI dataset.}
	\label{fig:4}
\end{figure}
We unify the input embedding and the position embedding of the typical transformer network into a local embedding module. The embedded local features for the template $F^{lt}=\left \{f_{i}^{lt}\right \}_{i=1}^{N_{1}}$ provide initial features for each point and unique local information since each point has unique coordinates.

\noindent\textbf{Target-aware feature learning.} Different from previous Siamese trackers, we develop the target-aware feature extraction and interaction method by a unified single-branch block and avoid the time-consuming feature fusing network, which is constructed based on the transformer network and self attention mechanism. After local feature embedding, the template local features $F^{lt}\in \mathbb{R}^{N_{1}\times 64}$ and search local features $F^{ls}\in \mathbb{R}^{N_{2}\times 64}$ are concatenated as $F^{l}=[F^{lt};F^{ls}]$. As shown in Fig. \ref{fig:2}, we feed $F^{l}$ into three stacked transformer blocks and obtain $F^{g}\in \mathbb{R}^{{(N_{1}+N_{2})}\times d}$. Then we split it to obtain target-aware search point features $F^{s}\in \mathbb{R}^{N_{2}\times d}$. The detailed structure of the transformer network is shown in Fig. \ref{fig:4}(a).

In each transformer block, we first use layer normalization operation to normalize the initial point features. Then, linear layers are utilized to map the inputs into query (Q), key (K), and value (V). Unlike the typical transformer structure, we abandon the special position encoding module since the coordinate-based local feature embedding network already provides unique position information for every point. The self attention mechanism is the core operation of our transformer network. It can be formulated as:

\begin{equation}
	attn=Softmax(\frac{Q K^{T}}{\sqrt d}) \label{eq5},
\end{equation}
\begin{equation}
	out= Softmax(\frac{Q K^{T}}{\sqrt d})\cdot V \label{eq6}
\end{equation}
where $attn$ is the attention map and $out$ is the output of the self attention mechanism. Especially, we concatenate the template and search point features as the input embeddings $F^{l}=[F^{lt};F^{ls}]$, thus the Q, K, V can be represented as $[Q^{t};Q^{s}]$, $[K^{t};K^{s}]$, and $[V^{t};V^{s}]$. Eq. \ref{eq5} and Eq. \ref{eq6} can be expanded as:
\begin{align}
	attn&=Softmax(\frac{[Q^{t};Q^{s}] [K^{t};K^{s}]^{T}}{\sqrt d})\notag\\
	&=[\lambda^{tt},\lambda^{ts};\lambda^{st},\lambda^{ss}]\label{eq7},
\end{align}
\begin{align}
	out&=[\lambda^{tt},\lambda^{ts};\lambda^{st},\lambda^{ss}]\cdot[V^{t};V^{s}]\notag\\
	&=[\lambda^{tt}V^{t}+\lambda^{ts}V^{s};\lambda^{st}V^{t}+\lambda^{ss}V^{s}]\label{eq8}
\end{align}
where $\lambda^{tt}$ and $\lambda^{ss}$ are attention weights, $\lambda^{ts}$ and $\lambda^{st}$ indicate the dual-interaction between template and search area points. Eq.\ref{eq7} and Eq.\ref{eq8} explain why the proposed single-branch backbone can learn target-aware point features for the search area. As shown in Fig. \ref{fig:4}(b), we visualize the attention maps generated by the above transformer network. We can find that the unified target-aware feature learning and interaction network can focus on the target well and distinguish the similar disturbances. It is worth noting that we apply the multi-head attention mechanism to enhance the representative ability of the network. Besides, the residual connection is considered and an MLP layer is designed to extract much deeper features.
\label{3.3}

\subsection{Efficient BEV-based Target Localization}
After feature learning, we obtain target-aware search point features $F^{s}\in \mathbb{R}^{N_{2}\times d}$. An efficient target localization is proposed to regress the target's location in the bird's eye view (BEV) feature space instead of directly regressing in sparse point clouds.
We first divide the irregular point clouds into equal voxels \cite{zhou2018voxelnet} according to their 3D coordinates. By performing max pooling along the z-axis on the voxelized feature maps, we obtain dense BEV feature maps where low responses in the voxelized feature maps can be suppressed. However, voxelized features are four-dimensional tensors and temporal fusion will superimpose features, making the encoder computationally intensive. A pragmatic approach involves the utilization of 3D convolutions applied to aggregate voxel features. We perform max pooling operation along the $z$ axis to project voxel features into the BEV feature space. However, this method is notably slow and inefficient. 

To address this issue, we propose an efficient encoder module (EEM) that reduces dimensionality with space-to-channel operations. Specifically, the module converts the 4D voxel tensor $V\in R^{H \times W \times Z \times C}$ into a 3D BEV tensor $V \in R^{H \times W \times (ZC)}$ via "\textit{torch. reshape}" operation, thus avoiding the need for memory-intensive 3D convolution. Given a target-aware search point features $F^{s}$ as input, we employ several 2D convolutions to reduce the channel dimension and incorporate residual connections to capture information from different layers. The module can be represented as follows:
 \begin{equation}
	F^v =Reshape(Voxel(F^s))
\end{equation}
\begin{equation}
	F^r = \alpha(F^v + \beta F^v)
\end{equation}
where $F^r\in \mathbb{R}^{H\times W \times d_{1}}$ represent the encoded features of the BEV feature. $\beta$ means two 3*3 2D convolutions and a ReLU layer. $\alpha$ represents two ReLU layers and a 3*3 2D convolution.

\begin{figure*}[t]
	\begin{center}
		\includegraphics[width=\textwidth]{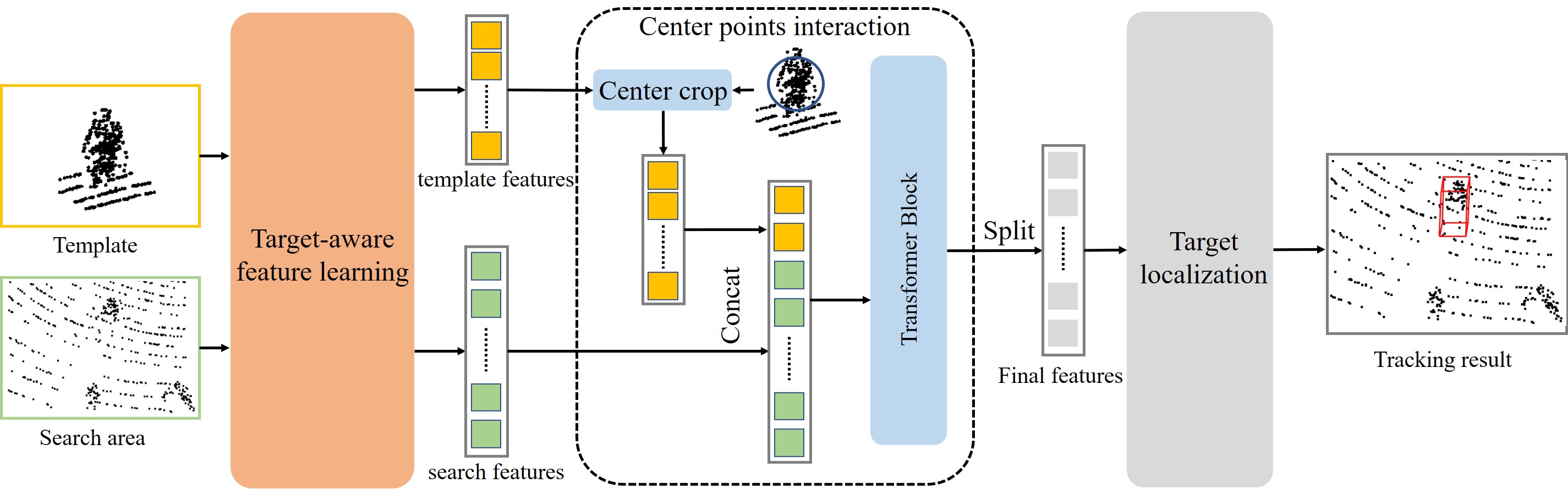}
	\end{center}
	\caption{The network structure of EasyTrack++. It is built on top of EasyTrack. Thus, the target-aware feature learning and target localization networks are the same as Fig. \ref{fig:2}. The center points interaction strategy crop the center points and then make secondary interaction with the search area points through the transformer network discussed in Fig. \ref{fig:4}.}
	\label{fig:5}
\end{figure*}
\begin{figure}[h]
	\begin{center}
		\includegraphics[width=\linewidth]{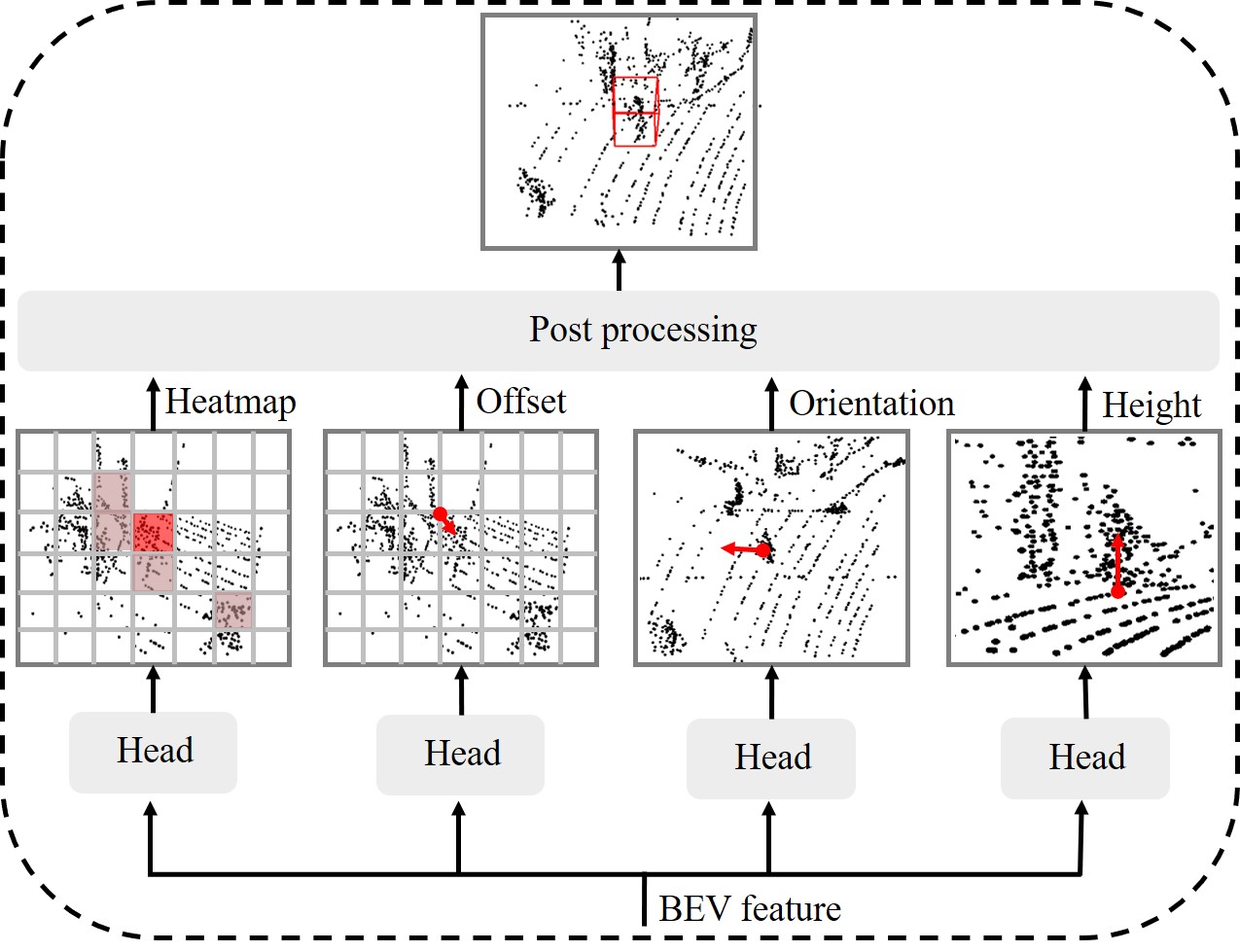}
		\caption{Illustration of the heads of the proposed EasyTrack. It consists of four parts to realize accurate classification and regression collaboratively.}
		\label{fig:6}
	\end{center}
\end{figure}
Inspired by \cite{ge2020afdet}, we design four heads from a decoupling perspective to achieve accurate tracking, as shown in Fig. \ref{fig:5}. A heatmap $M\in \mathbb{R}^{H\times W \times 1}$ is predicted to find which pixel in the BEV space in the target's center is located. The offset map $O\in \mathbb{R}^{H\times W \times 2}$ helps find a more accurate center and recover the discretization error. Orientation $\Theta\in \mathbb{R}^{H\times W \times 1}$ determines the heading angle of the target. We also predict the height of the target's center $Z\in \mathbb{R}^{H\times W \times 1}$ to locate the target in the $z$-axis. After post processing, we can decode the target location $(x,y,z,w,l,h,\theta)$ in the search area.

The total loss function of our method can be formulated as:
\begin{equation}
	L=\lambda_{1}L_{cls} + \lambda_{2}L_{off} + \lambda_{3}L_{\theta} + \lambda_{4}L_{Z}
\end{equation}
Among them, $L_{cls}$ is a modified focal loss for heatmap $M$, $L_{off}$ is a $L_{1}$ loss for offset map $O$, $L_{\theta}$ and $L_{Z}$ are also $L_{1}$ losses for Orientation $\Theta$ and height $Z$ respectively. $\lambda_{1},\lambda_{2},\lambda_{3},\lambda_{4}$ are weights for different parts.
\label{sec:3.3}

\subsection{Center points interaction strategy}
Although EasyTrack has demonstrated superior performance and achieved a balance between tracking accuracy and speed, we still make improvements to it and propose an enhanced version named EasyTrack++. In 3D single object tracking task, the template usually contains information around the target. This character brings heavy interaction noise when matching template and search area. To address this problem, SimTrack\cite{chen2022backbone} propose a foveal window strategy to provide more diversified template information in 2D SOT. Here, we develop a center points interaction strategy that specially designed for 3D SOT in point clouds.

\begin{figure*}[t]
	\begin{center}
		\includegraphics[width=\textwidth]{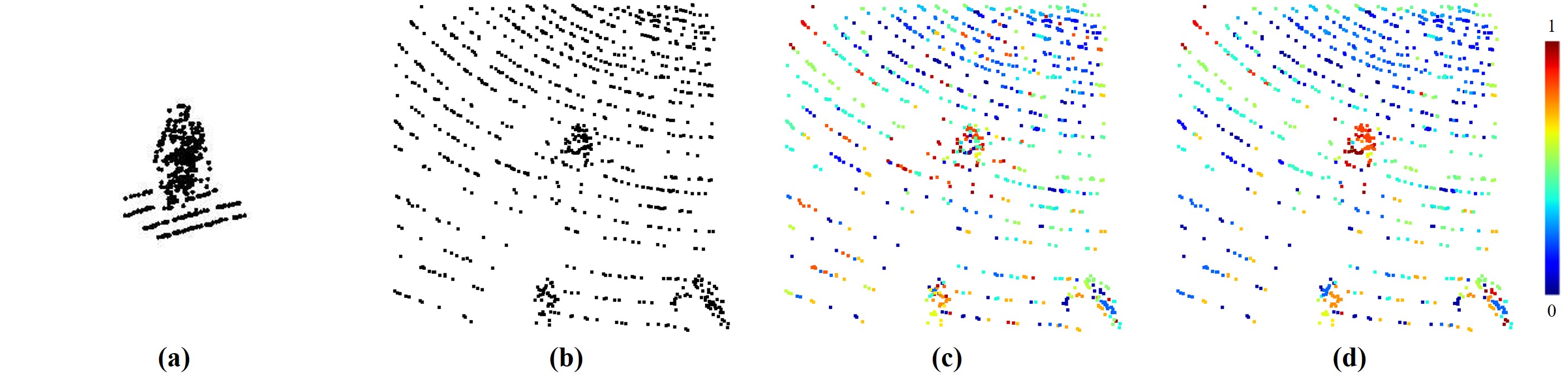}
	\end{center}
	\caption{The visualization comparison of attention maps generated by the global interactions in EasyTrack and EasyTrack++. We select one tracking scenes in the Pedestrian category in the KITTI dataset. (a) the template points. (b) the search area points. (c) the attention map generated by EasyTrack. (d) the attention map generated by EasyTrack++.}
	\label{fig:7}
\end{figure*}

As shown in Fig. \ref{fig:7}(a), the template contains points belonging to pedestrians located in the center and background points distributed around them. We utilize a simple but effective cropping strategy to crop the center points to make secondary interactions with the search points to emphasize target information in the search area. Specifically, after target-aware feature learning, we obtain template features $F^{t}\in \mathbb{R}^{N_{1}\times d}$ and search area features $F^{s}\in \mathbb{R}^{N_{2}\times d}$. We first use the simplified ball query to define a central area around the geometric center $o$ of the template $P^{t}\in \mathbb{R}^{N_{1}\times 3}$. Then we collect a fixed number of points $P^{c}\in \mathbb{R}^{N_{3}\times 3}$ inside it and aggregate their features in $F^{t}$ as $F^{c}\in \mathbb{R}^{N_{3}\times d}$, where $N_{3}$ is the fixed number. If the number of points inside the central area is more than $N_{3}$, we use random sampling to sample $N_{3}$ points. Otherwise, we repeat the existing points. After that, we concatenate features of center points and search area and make secondary interactions through the transformer block mentioned in Sec~\ref{3.3}. The whole center points interaction strategy can be formulated as:
 \begin{equation}
 	(P^{c},F^{c})=BallQuery(P^{t},F^{t}, o)
 \end{equation}
 \begin{equation}
 	F^{f}=Transformer(concat(F^{c},F^{s}))
 \end{equation}
where $F^{f}$ is the final feature for target localization. We visualize the attention maps generated by the proposed EasyTrack and EasyTrack++ in Fig. \ref{fig:7}. As shown in Fig. \ref{fig:7}(c), in EasyTrack, some background points around the target have high response that maybe confuse the target localization network and lead to inaccurate classification and regression. However, as shown in Fig. \ref{fig:7}(d), the response of target is much more distinct than other backgrounds. This further reflects that the center points interaction strategy enhances the discrimination of feature extraction networks.

\begin{table*}[t]
	\centering
	\caption{Tracking results compared to other trackers in the KITTI dataset. Success / Precision are used for evaluation. \textbf{Bold} denotes the best result.}
	\resizebox{178mm}{43mm}{
		\begin{tabular}{c|ccccc|ccccc}
			\hline
			Metric & \multicolumn{5}{c|}{Success} & \multicolumn{5}{c}{Precision} \bigstrut\\
			\hline
			Category & Car & Pedestrian & Van & Cyclist & Mean & Car & Pedestrian & Van & Cyclist & Mean \bigstrut[t]\\
			Frame Number & 6424  & 6088  & 1248  & 308  & 14068  & 6424  & 6088  & 1248  & 308  & 14068 \bigstrut[b]\\
			\hline
			SC3D\cite{giancola2019leveraging} & 41.3  & 18.2  & 40.4  & 41.5  & 31.2  & 57.9  & 37.8  & 47.0  & 70.4  & 48.5  \bigstrut[t]\\
			P2B\cite{qi2020p2b} & 56.2  & 28.7  & 40.8  & 32.1  & 42.4  & 72.8  & 49.6  & 48.4  & 44.7  & 60.0  \\
			MLVSNet\cite{wang2021mlvsnet} & 56.0  & 34.1  & 52.0  & 34.3  & 45.7  & 74.0  & 61.1  & 61.4  & 44.5  & 66.7  \\
			LTTR\cite{lttr} & 65.0  & 33.2  & 35.8  & 66.2  & 48.7  & 77.1  & 56.8  & 45.6  & 89.9  & 65.8  \\
			BAT\cite{zheng2021box} & 60.5  & 42.1  & 52.4  & 33.7  & 51.2  & 77.7  & 70.1  & 67.0  & 45.4  & 72.8  \\
			PTT\cite{shan2021ptt} & 67.8  & 44.9  & 43.6  & 37.2  & 55.1  & 81.8  & 72.0  & 52.5  & 47.3  & 74.2  \\
			C2FT\cite{fan2022accurate} & 67.0  & 48.6  & 53.4  & 38.0  & 57.2  & 80.4  & 75.6  & 66.1  & 48.7  & 76.4  \\
			V2B\cite{hui20213d} & 70.5  & 48.3  & 50.1  & 40.8  & 58.4  & 81.3  & 73.5  & 58.0  & 49.7  & 75.2  \\
			PTTR\cite{zhou2022pttr} & 65.2  & 50.9  & 52.5  & 65.1  & 57.9  & 77.4  & 81.6  & 61.8  & 90.5  & 78.1  \\
			CMT\cite{guo2022cmt} & 70.5  & 49.1  & 54.1  & 55.1  & 59.4  & 81.9  & 75.5  & 64.1  & 82.4  & 77.6  \\
			STNet\cite{hui20223d} & 72.1  & 49.9  & 58.0  & 73.5  & 61.3  & 84.0  & 77.2  & 70.6  & 93.7  & 80.1  \\
			$M^{2}$-Track\cite{zheng2022beyond} & 65.5  & 61.5  & 53.8  & 73.2  & 62.9  & 80.8  & 88.2  & 70.7  & 93.5  & 83.4\\
			STTrack\cite{cui2023sttracker} & 66.5  & 60.4  & 50.5  & 75.3  & 62.6  & 79.9  & 89.4  & 63.6  & 93.9  & 82.9  \\
			PCET\cite{wang2023implicit} & 68.7  & 56.9  & 57.9  & 75.6  & 62.7  & 80.1  & 85.1  & 66.1  & 93.7  & 81.3 \\
			PTIT\cite{liu2023instance} & 68.6  & 56.7  & 53.8  & 74.2  & 62.6  & 81.2  & 86.3  & 70.7  & 92.5  & 82.7 \\
			CXTrack\cite{xu2023cxtrack} & 69.1  & 67.0  & 60.0  & 74.2  & 67.5  & 81.6  & 91.5  & 71.8  & 94.3  & 85.3  \\
			CorpNet\cite{wang2023correlation} & 73.6  & 74.3  & 58.7  & 55.6  & 64.5  & 84.1  & 94.2  & 66.5  & 82.4  & 82.0\\
			SyncTrack\cite{ma2023synchronize} & 73.3  & 54.7  & 60.3  & 73.1  & 64.1  & 85.0 & 80.5  & 70.0  & 93.8  & 81.9 \\
			MBPTrack\cite{xu2023mbptrack} & 73.4  & 68.6  & 61.3  & 76.7  & 70.3  & 84.8  & 93.9  & 72.7  & 94.3  & 87.9   \\
			M3SOT\cite{liu2023m3sot} & 75.9  & 66.6  & 59.4  & 70.3  & 70.3  & 87.4  & 92.5  & 74.7  & 93.4  & 88.6    \bigstrut[b]\\
			\hline
			EasyTrack(ours) & 87.9  & 88.2  & 65.3  & 85.3  & 86.0  & 90.1  & 95.6  & 78.8  & 93.8  & 91.6  \bigstrut[t]\\
			EasyTrack++(ours) & \textbf{88.7}  & \textbf{91.2}  & \textbf{68.6} & \textbf{87.4}  & \textbf{88.0}  & \textbf{92.6}  & \textbf{96.3}  & \textbf{80.0}  & \textbf{95.4}  & \textbf{93.2}  \bigstrut[b]\\
			\hline
	\end{tabular}}%
	\label{tab:1}%
\end{table*}%
\begin{table*}[t]
	\centering
	\small
	\caption{Tracking results compared to other trackers in the nuScenes dataset. Success / Precision are used for evaluation. \textbf{Bold} denotes the best result.}
	\resizebox{178mm}{22mm}{
		\begin{tabular}{c|ccccc|ccccc}
			\hline
			Category & Car & Pedestrian & Truck & Bicycle & Mean & Car & Pedestrian & Truck & Bicycle & Mean \bigstrut[t]\\
			Frame Number & 15578  & 8019  & 3710  & 501  & 27808  & 15578  & 8019  & 3710  & 501  & 27808  \bigstrut[b]\\
			\hline
			SC3D\cite{giancola2019leveraging} & 25.0  & 14.2  & 25.7  & 17.0  & 21.8  & 27.1  & 16.2  & 21.9  & 18.2  & 23.1  \bigstrut[t]\\
			P2B\cite{qi2020p2b} & 27.0  & 15.9  & 21.5  & 20.0  & 22.9  & 29.2  & 22.0  & 16.2  & 26.4  & 25.3  \\
			BAT\cite{zheng2021box} & 22.5  & 17.3  & 19.3  & 17.0  & 20.5  & 24.1  & 24.5  & 15.8  & 18.8 & 23.0  \\
			V2B\cite{hui20213d} & 31.3  & 17.3  & 21.7  & 22.2  & 25.8  & 35.1  & 23.4  & 16.7  & 19.1  & 29.0  \\
			STNet\cite{hui20223d} & 32.2  & 20.4  & 27.6  & 18.5  & 26.5  & 33.4  & 32.9  & 20.8  & 26.8  & 31.5 \\	
			CXTrack\cite{xu2023cxtrack} & 29.6  & 20.4  & 27.6  & 18.5  & 26.5  & 33.4  & 32.9  & 20.8  & 26.8  & 31.5  \\
			CorpNet\cite{wang2023correlation} & 35.0  & 21.3  & 39.7  & 26.9  & 31.8  & 38.4  & 33.6  & 36.3  & 43.5  & 36.8   \\
			SyncTrack\cite{ma2023synchronize} & 36.7  & 19.1  & 39.4  & 23.8  & 31.8  & 38.1 & 27.8  & 38.6  & 30.4  & 35.1 \\
			M3SOT\cite{liu2023m3sot} & 34.2  & 24.6  & 29.6  & 18.8  & 30.5  & 38.6 & 37.8  & 25.5  & 27.9  & 36.4 \bigstrut[b] \\
			\hline
			EasyTrack(ours) & 76.1  & 60.5  & 71.8  & 44.8  & 70.5  & 80.5  & 86.9  & 70.0  & 73.2  & 80.8  \bigstrut[t]\\
			EasyTrack++(ours) & \textbf{76.8}  & \textbf{61.4} & \textbf{72.1}  & \textbf{45.1}  & \textbf{71.2}  & \textbf{80.9}  & \textbf{87.2}  & \textbf{70.2}  & \textbf{74.5}  & \textbf{81.2}  \bigstrut[b]\\
			\hline
	\end{tabular}}%
	\label{tab:2}%
\end{table*}%
\begin{table*}[htbp]
	\centering
	\small
	\caption{Tracking results compared to other trackers in the WOD. Success / Precision are used for evaluation. \textbf{Bold} denotes the best result.}
	\resizebox{178mm}{21mm}{
	\begin{tabular}{c|cccc|cccc|c}
		\hline
		\multirow{2}[2]{*}{Category} & \multicolumn{4}{c|}{Vehicle(185731)} & \multicolumn{4}{c|}{Pedestrian(241752)} & \multirow{2}[2]{*}{Mean} \bigstrut[t]\\
		& Easy  & Medium & Hard  & Mean  & Easy  & Medium & Hard  & Mean  &  \bigstrut[b]\\
		\hline
		P2B\cite{qi2020p2b}   & 57.1/65.4 & 52.0/60.7 & 47.9/58.5 & 52.6/61.7 & 18.1/30.8 & 17.8/30.0 & 17.7/29.3 & 17.9/30.1 & 33.0/43.8 \bigstrut[t]\\
		BAT\cite{zheng2021box}   & 61.0/68.3 & 53.3/60.9 & 48.9/57.8 & 54.7/62.7 & 19.3/32.6 & 17.8/29.8 & 17.2/28.3 & 18.2/30.3 & 34.1/44.4 \\
		V2B\cite{hui20213d}   & 64.5/71.5 & 55.1/63.2 & 52.0/62.0 & 57.6/65.9 & 27.9/43.9 & 22.5/36.2 & 20.1/33.1 & 23.7/37.9 & 38.4/50.1 \\
		TAT\cite{lan2022temporal}   & 66.0/72.6 & 56.6/64.2 & 52.9/62.5 & 58.9/66.7 & 32.1/49.5 & 25.6/40.3 & 21.8/35.9 & 26.7/42.2 & 40.7/52.8 \\
		STNet\cite{hui20223d} & 65.9/72.7 & 57.5/66.0 & 54.6/64.7 & 59.7/68.0 & 29.2/45.3 & 24.7/38.2 & 22.2/35.8 & 25.5/39.9 & 40.4/52.1 \\
		CXTrack\cite{xu2023cxtrack} & 63.9/71.1 & 54.2/62.7 & 52.1/63.7 & 57.1/66.1 & 35.4/55.3 & 29.7/47.9 & 26.3/44.4 & 30.7/49.4 & 42.2/56.7 \\
		$M^{2}$-Track\cite{zheng2022beyond} & 68.1/75.3 & 58.6/66.6 & 55.4/64.9 & 61.1/69.3 & 35.5/54.2 & 30.7/48.4 & 29.3/45.9 & 32.0/49.7 & 44.6/58.2 \\
		MBPTrack\cite{xu2023mbptrack} & 68.5/77.1 & 58.4/68.1 & 57.6/69.7 & 61.9/71.9 & 37.5/57.0 & 33.0/51.9 & 30.0/48.8 & 33.7/52.7 & 46.0/61.0 \bigstrut[b]\\
		\hline
		EasyTrack++(Ours)  & \textbf{70.0}/\textbf{77.8} & \textbf{59.1}/\textbf{69.1} & \textbf{58.3}/\textbf{70.5} & \textbf{62.8}/\textbf{72.7} & \textbf{38.1}/\textbf{58.5} & \textbf{35.2}/\textbf{52.2} & \textbf{31.5}/\textbf{49.0} & \textbf{35.1}/\textbf{53.5} & \textbf{47.1}/\textbf{61.8} \bigstrut\\
		\hline
	\end{tabular}}%
	\label{tab:3}%
\end{table*}%
\begin{table}[htbp]
	\centering
	\Large
	\caption{The results for the actual tracked frames in the nuScenes dataset.}
	\resizebox{\linewidth}{!}{
		\begin{tabular}{cccccc}
			\toprule
			Category & Car   & Pedestrian & Truck & Bicycle & Mean \\
			Frame Number & 146871 & 75066 & 35448 & 4698  & 262123 \\
			\midrule
			V2B\cite{hui20213d}  &34.5/36.1 & 18.9/23.6 & 32.9/29.3 & 23.2/30.5 & 29.6/31.5  \\
			Easytrack++ & \textbf{74.5}/\textbf{78.8} & \textbf{58.5}/\textbf{84.7} & \textbf{70.1}/\textbf{68.3} & \textbf{38.5}/\textbf{72.1} & \textbf{68.7}/\textbf{78.9} \\
			\bottomrule
		\end{tabular}%
	}
	\label{tab:addlabel3}%
\end{table}%

\section{Experiments}
In this section, we conduct extensive experiments to show the favorable performance of EasyTrack and EasyTrack++. First, we introduce our experiment settings including datasets and evaluation metrics. Then we give comprehensive comparisons with other state-of-the-art trackers to validate the superiority of our tracker. Last, adequate ablation studies validate the effectiveness of each part in the EasyTrack and EasyTrack++.


\subsection{Experiment settings}
\noindent\textbf{Datasets.} We evaluate the performance of EasyTrack and EasyTrack++ on the challenging KITTI\cite{geiger2012we}, nuScenes\cite{caesar2020nuscenes} and Waymo Open Dataset (WOD)\cite{sun2020scalability}. They mainly focus on autonomous driving scenes. KITTI has 21 training scenes and we follow \cite{giancola2019leveraging} to split the training set since the testing set has no annotations. Specifically, we train in scenes 0-16, validate in scenes 17-18, and test in scenes 19-20. For the nuScenes dataset, we follow \cite{hui20213d} to split it into 750 training sequences and 150 validation sequences. We train our method on the training set and test it on its validation set. Notably, we use the official toolkit to insert ground truth for unannotated frames since only key frames have annotations in nuScenes. For WOD, we followed LiDAR SOT\cite{pang2021model} to evaluate our approach on 1,121 tracks divided into easy, medium, and difficult subsets based on point cloud sparsity. The datasets use for our pre-training include ShapeNet,ModelNet40,ScanObjectNN. The ShapeNet\cite{chang2015shapenet} dataset utilized for pre-training comprises approximately 51,300 meticulously curated 3D models, encompassing 55 prevalent object categories. It was meticulously compiled by collecting CAD models from open-source 3D repositories available online. CAD models, which are digital three-dimensional representations within computer-aided design software, exhibit highly precise geometric information. ModelNet40 is a commonly used dataset for 3D object recognition that contains 40 different categories of 3D CAD models. The models were divided into 12,311 training samples and 1,269 test samples. The object category in ModelNet40 covers common objects such as chairs, tables, monitors, airplanes, and so on. Unlike the first two, ScanObjectNN is a challenging real-world dataset consisting of approximately 15,000 objects from 15 categories that provide more light variation, noise interference, and target context information to simulate complex and variable real-world scenarios.

\noindent\textbf{Evaluation Metrics.} We follow previous methods to report the Success and Precision metrics. Among them, the Success metric is defined by the 3D IOU between the predicted 3D bounding box and the ground truth. The Precision metric is defined by the Area Under Curve (AUC) for the distance between centers of the predicted and annotated bounding boxes from 0 to 2m.

\noindent\textbf{Model Details.} In the proposed EasyTrack and EasyTrack++, we take $N_{1}=512$ template points and $N_{2}=1024$ search area points as input. The radius $r = 0.3m$ and $K=32$ neighbor local points are considered in the local feature embedding network. In the target-aware feature learning process, we use the multi-head self attention in the transformer block. We stack three transformer blocks with four heads. The feature dimension $d$ is set to 64. In the center points interaction stage, we collect $N_{3}=128$ points to form $P^{c}$ and we set use one layer of transformer network for secondary interaction. In the target localization network, the voxel size is set to $(0.3m,0.3m,0.3m)$. The size of the BEV feature map is $24\times38\times128$. Four 2D convolutions and two 2D deconvolutions are designed to process the BEV feature in combination with the residual connection.

\noindent\textbf{Training.} In the training stage, we merge point clouds in the $(t-1)$-th bounding box and the first bounding box as the template. The point clouds inside the enlarged $t$-th bounding box are sampled as the search area along with random shift. We set $\lambda_{1} = 1.0,\lambda_{2}=1.0,\lambda_{3}=1.0,\lambda_{4}=2.0$ in the loss function. Our model is end-to-end trained for 20 epochs. We use the Adam optimizer, and the initial learning rate is set to 0.001. It is divided by 5 every 6 epochs.

\noindent\textbf{Testing.} In the testing stage, we give the target in the first frame of a tracklet and track it across all frames. We merge point clouds in the $(t-1)$-th predicted result and the given bounding box in the first frame as the template. The $(t-1)$-th predicted result is enlarged by $2m$ in the $t$-th frame and we sample 1024 points inside it as the search area.

\noindent\textbf{3D Pre-training.} Empirical evidence reveals that pre-trained models on 3D analysis task datasets (e.g. ShapeNet dataset) can significantly enhance the precision of 3D tracking tasks. The ShapeNet dataset comprises approximately 50,000 3D CAD models spanning 14 major categories and 55 subcategories. To be more specific, our proposed target-aware feature learning network serves as the backbone, and we pre-trained it on ShapeNet for 300 epochs using Point-MAE. During training on the KITTI and nuScenes datasets, we initialized EasyTrack with pre-trained weights from the backbone network of the 3D analysis framework. The remaining layers were initialized randomly.

\begin{figure}[t]
	\begin{center}
		\includegraphics[width=\linewidth]{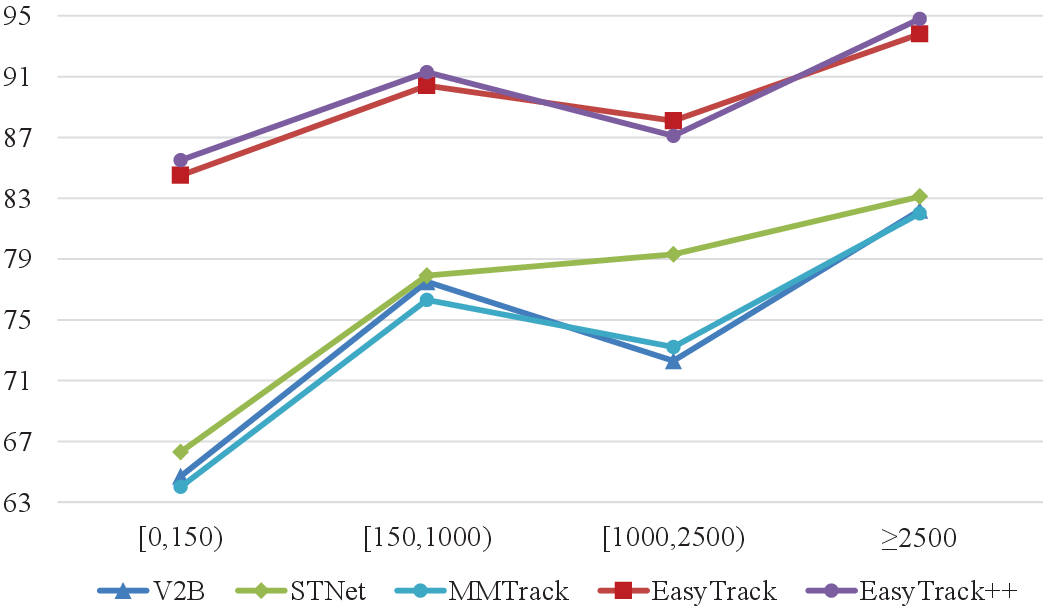}
	\end{center}
	\caption{Robustness analysis in the Car category in the KITTI dataset.}
	\label{fig_zhexian}
\end{figure}
\subsection{Comparison with other trackers}
\noindent\textbf{Results on KITTI.} We compare EasyTrack and EasyTrack++ with other state-of-the-art 3D SOT methods in Table \ref{tab:1}. Following previous works, we report the results in four categories including Car, Pedestrian, Van, and Cyclist. The mean results are calculated based on all frames. It can be observed that the proposed EasyTrack outperforms other trackers across all categories, with an average success and precision score of 86.0/91.6. Besides, thanks to the center points interaction strategy, the enhanced version EasyTrack++ further improves the tracking performance. The mean Success and Precision metrics are the highest among all trackers (88.0/93.2). Compared to EasyTrack, EasyTrack++ shows an improvement of 2.0 in mean Success and 1.6 in Precision metrics compared to EasyTrack.

\noindent\textbf{Results on nuScenes.} We show the tracking results on nuScenes dataset compared to other state-of-the-art trackers in Table \ref{tab:2}. It is important to note that for unannotated frames, the ground truth is obtained through interpolation, as mentioned earlier. We compare with those trackers that follow the same settings. We report the results in four categories including Car, Pedestrian, Truck, and Bicycle. Note that the results are computed on the key frames of the validation set. EasyTrack outperforms other trackers in all categories. EasyTrack++ further improves the tracking performance and achieves the best mean Success and precision (71.2/81.2). Compared to tracking in the KITTI dataset, it is more difficult to track the target in nuScenes due to the low-quality interpolated ground truth. In previous work, tracking results were reported only for keyframes (Car-15578), as shown in Table \ref{tab:2}. However, in Table \ref{tab:addlabel3}, we provide results for all actual tracked frames (Car-145871), and EasyTrack++ exhibits a significant advantage over V2B.

\begin{figure*}[t]
	\begin{center}
		\includegraphics[width=\textwidth]{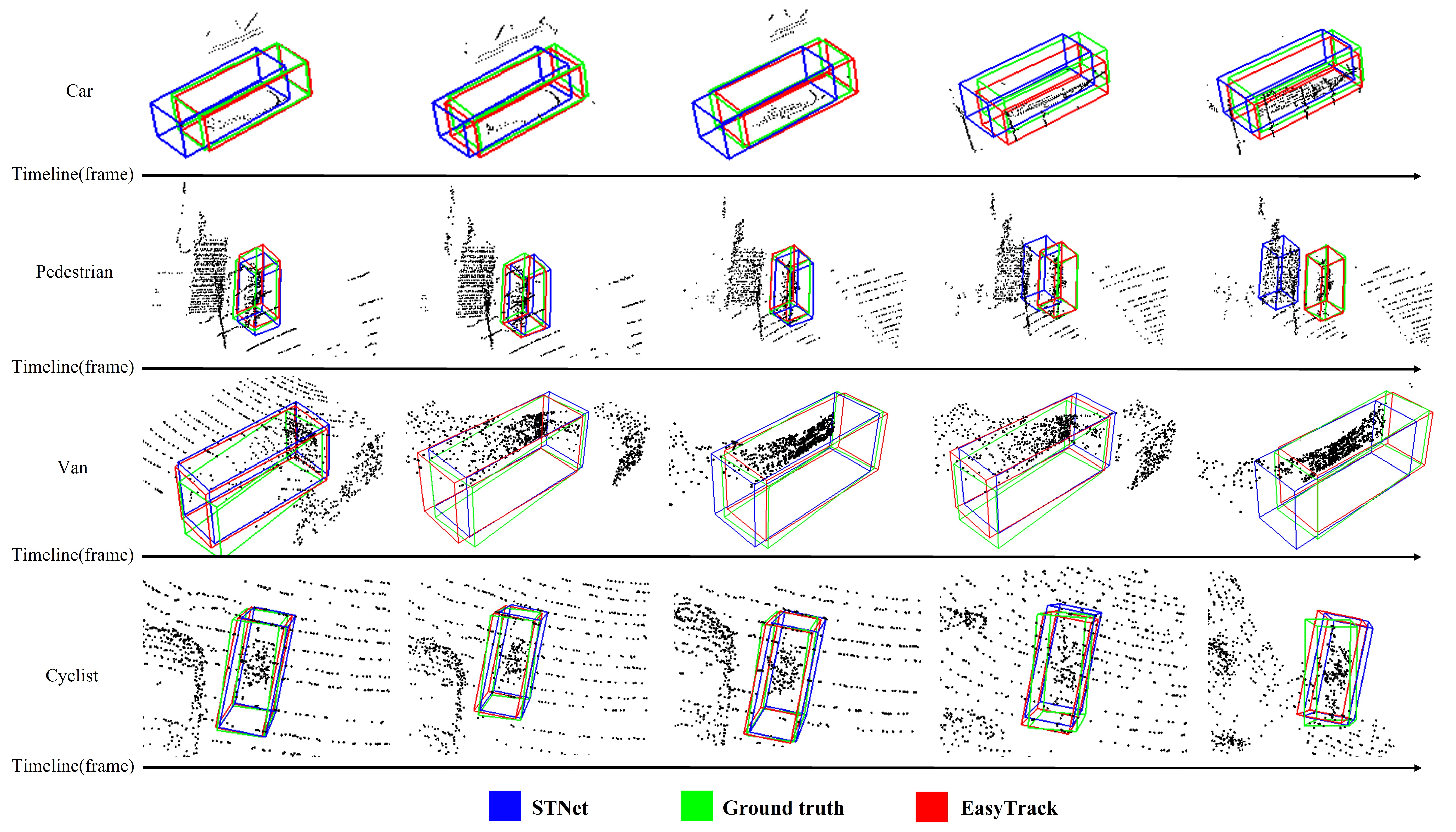}
	\end{center}
	\caption{Visualization of advantage cases compared to STNet\cite{hui20223d}. We visualize one tracking scene for each category in the KITTI dataset. }
	\label{fig:9}
\end{figure*}
\begin{figure*}[t]
	\begin{center}
		\includegraphics[width=0.985\textwidth]{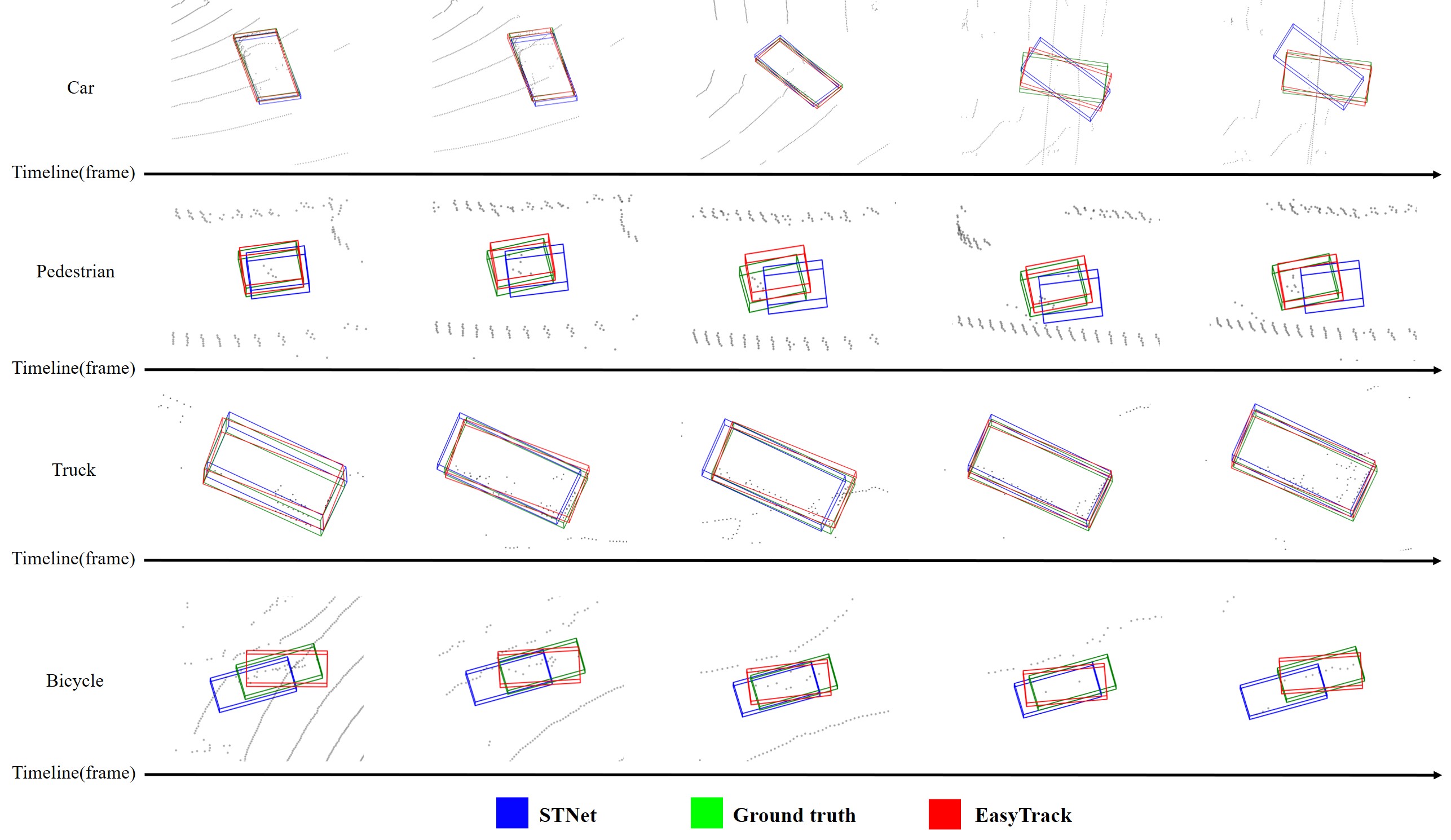}
	\end{center}
	\caption{Visualization of advantage cases compared to STNet\cite{hui20223d}. We visualize one tracking scene for each category in the nuScenes dataset. }
	\label{fig:10}
\end{figure*}
\noindent\textbf{Results on Waymo.} To assess the generalization capabilities of our proposed EasyTrack++, we conducted an evaluation of the KITTI pre-trained model on the WOD, as outlined in the prior work \cite{hui20223d}. It is worth noting that the object categories between KITTI and WOD align as follows: Car $\rightarrow$ Vehicles and Pedestrian $\rightarrow$ Pedestrian. The experimental results, presented in Table \ref{tab:3}, Comparison with other methods reveal that EasyTrack++ exhibits superior tracking performance across various challenging occlusion scenarios, achieving the highest average success rate and precision (47.1/61.8). In summary, our proposed methodology not only demonstrates accurate target tracking across diverse object sizes (vehicle or pedestrian) but also exhibits strong generalization capabilities when applied to previously unobserved scenes.

\begin{table}[t]
	\centering
	\caption{Computation costs compared to other Siamese trackers.}
	\vspace{2mm}
	\begin{tabular}{cccc}
		\hline
		Method & Param(M) & FLOPs(G) & FPS \bigstrut\\
		\hline
		P2B\cite{qi2020p2b} & 1.3 & 4.7 & 35.7 \bigstrut[t]\\
		BAT\cite{zheng2021box} & 1.5 & 3.1 & 51.8 \\
		PTT\cite{shan2021ptt} & 4.9 & 6.2 & 41.2 \\
		V2B\cite{hui20213d} & 1.3 & 5.6 & 20.4 \\
		PTTR\cite{zhou2022pttr} & 2.1 & 2.7 & 35.9 \\
		STNet\cite{hui20223d} & 2.0 & 3.1 & 27.5 \\
		$M^{2}$-Track\cite{zheng2022beyond} & 2.2 & 2.5 & 57.0 \\
		EasyTrack& \textbf{1.0} & \textbf{2.3} & \textbf{53.2} \\
		EasyTrack++& 1.3 & 2.6 & 52.6 \bigstrut[b]\\
		\hline
	\end{tabular}%
	\label{tab:4}%
\end{table}%

\noindent\textbf{Computational cost.} We propose a simple baseline for 3D SOT without the heavy feature fusion module that is necessary for typical Siamese trackers. The computational costs of different trackers are reported in Table \ref{tab:4}. For a fair comparison, we use a single NVIDIA 3090 GPU to test all the trackers. Table \ref{tab:4} reflects that EasyTrack has fewer parameters and FLOPs than other trackers. The proposed object localization module effectively reduces the model's parameter training volume with the compromising accuracy. It runs at a real-time speed (53.2FPS) and is only slow than $M^{2}$-Track but much more accurate as shown in Table \ref{tab:1}. Compared with typical Siamese trackers, EasyTrack is a lightweight tracker and achieves a good balance between running speed and tracking accuracy. At the same time, EasyTrack++ further improves tracking performance with the acceptable computation cost. It achieves the best tracking performance as illustrated in Tab~\ref{tab:1} and still outperforms most 3D SOT trackers in the speed metric and runs in a 52.6FPS.

\begin{figure*}[htbp]
	\begin{center}
		\includegraphics[width=\textwidth]{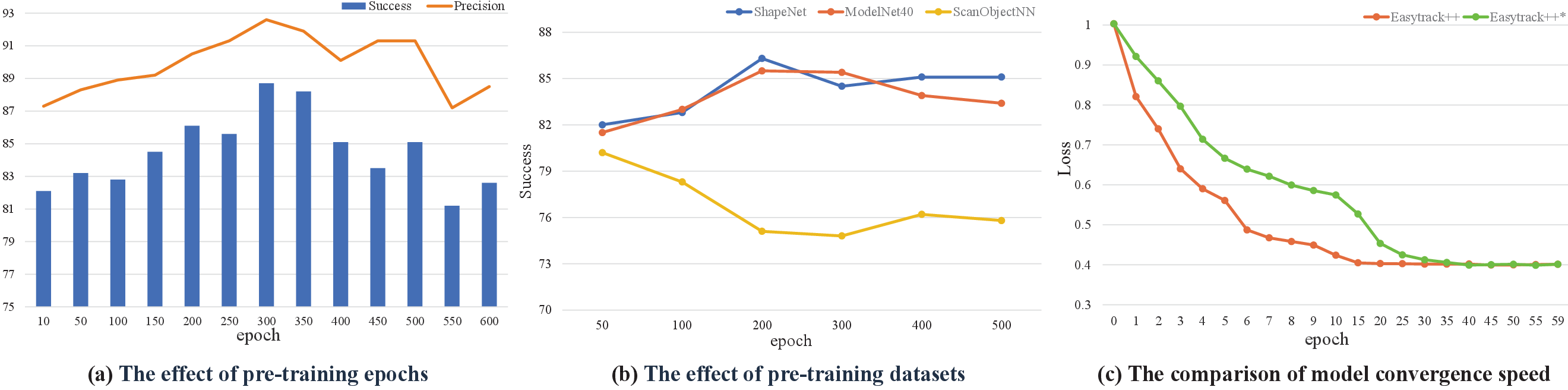}
	\end{center}
	\caption{(a) The impact of different pre-training epochs on tracking performance in the KITTI dataset. (b) The effect of pre-training datasets. (c) The convergence speed of the model in the KITTI dataset. $*$ indicates no pre-training employed.}
	\label{fig:loss_pre}
\end{figure*}
\noindent\textbf{Robustness in Sparse Scenes.} It is crucial for a 3D tracker to perform robust tracking in sparse point clouds for practical applications. We divide the tracing scenes into four intervals according to the number of points. We report the Success metric in the Car category in the KITTI dataset in Fig. \ref{fig_zhexian}. As the number of points increases, the performance of trackers increases steadily. EasyTrack presents better performance than V2B\cite{hui20213d} and STNet\cite{hui20223d} in three intervals. It reflects that EasyTrack has good robustness in sparse scenes that have incomplete target information. The proposed EasyTrack++ gradually improves the performance of EasyTrack since the center points interaction strategy has more informative points to guide the secondary interaction in relatively dense scenes. Fig. \ref{fig:loss_pre}(b) intuitively reflects that EasyTrack and EasyTrack++ have good robustness in sparse scenes. This is critical when facing occlusion conditions or tracking distant objects.

\noindent\textbf{Visualization comparisons.} To demonstrate the superiority of the proposed EasyTrack more intuitively, we visualize the tracking process compared to the competitive tracker STNet\cite{hui20223d} in Fig. \ref{fig:9} and Fig. \ref{fig:10}. In the Car category in the KITTI dataset, as shown in the first row of Fig. \ref{fig:9}, the point clouds of the target are quite sparse and incomplete. EasyTrack tracks well while STNet deviates largely. It reflects the good robustness of EasyTrack in sparse scenes. In a scene with more disturbs in the second row, EasyTrack holds the target pedestrian tightly while STNet gradually loses the target and locates a similar target. This can be attributed to the superiority of our target-aware feature learning process. The simple correlation operation in STNet leads to the unreliable feature matching among incomplete point clouds. Compared to KITTI dataset that uses a 64-line LIDAR, nuScenes dataset uses a 32-line LiDAR, thus the points are usually more sparse. As shown in Fig. \ref{fig:10}, EasyTrack also performs better than STNet in the sparse tracking scenes.


\subsection{Ablation study}
For a more thorough assessment of the individual efficacy of the proposed method, it is imperative to note that, unless otherwise specified, ablation experiments were conducted without the utilization of the pre-training strategy.

\noindent\textbf{Model components.} Our ablation study examined the impact of 3D SOT pre-training (Pre), target-aware feature learning (TA), and target localization (TL) during the training process. The results are shown in Table \ref{tab:addlabe2}. It was observed that the pre-training strategy significantly enhances tracking performance (\textcolor{green!70}{$\uparrow$21.2}/\textcolor{green!70}{$\uparrow$9.4}). This finding substantiates the superiority of the pre-training strategy specifically tailored for 3D SOT tasks, emphasizing the complementary value of the CAD model-based ShapeNet dataset used in pre-training for point cloud data. Even without employing the pre-training strategy, our tracker still demonstrates competitive results, with a mean success and precision of 64.5/83.5. Furthermore, our proposed target-aware feature learning and object localization methods exhibited varying degrees of performance improvement. Subsequently, with the implementation of the proposed target-aware feature learning, the mean performance is enhanced by 1.0/0.7. Finally, by leveraging the proposed target localization, we achieve a further performance boost of 1.2/0.5. This demonstrates its ability to accurately localize objects in the fusion space based on BEV features.
\begin{table}[t]
	\centering
	\Large
	\caption{Ablation studies on different model components on KITTI\cite{geiger2012we} dataset.}
	\resizebox{\linewidth}{!}{
		\begin{tabular}{ccccccccc}
			\toprule
			&Pre   & TA    & TL    & Car   & Pedestrian & Van   & Cyclist & Mean \\
			\midrule
			& {\footnotesize \XSolidBrush}     & {\footnotesize \XSolidBrush}     & {\footnotesize \XSolidBrush}     & 71.5/82.0 & 53.8/82.3 & 56.9/65.9 & 72.7/\textbf{93.8} & 62.6/81.0 \\
			& {\footnotesize \XSolidBrush}     & \checkmark     & \checkmark     & 72.8/84.0 & 56.5/85.8 & 58.6/67.5 & 73.1/93.7 & 64.5/83.5 \\
			&\checkmark     &{\footnotesize \XSolidBrush}     & {\footnotesize \XSolidBrush}     & 85.2/88.5 & 87.1/95.0 & 60.1/76.5 & 83.2/93.7 &83.8/90.4  \\
			&\checkmark     & \checkmark     & {\footnotesize \XSolidBrush}    & 86.5/89.6 & 87.9/95.2 & 60.8/77.5 & 84.3/93.7 &84.8/91.1  \\
			&\checkmark     & \checkmark     & \checkmark     & \textbf{87.9}/\textbf{90.1} & \textbf{88.2}/\textbf{95.6} & \textbf{65.3}/\textbf{78.8} & \textbf{85.3}/\textbf{93.8} &\textbf{86.0}/\textbf{91.6}  \\
			\bottomrule
		\end{tabular}%
	}
	\label{tab:addlabe2}%
\end{table}%

\begin{table}[t]
	\centering
	\tiny
	\caption{Results of different target localization networks. }
	\resizebox{\linewidth}{8mm}{
	\begin{tabular}{ccccc}
		\hline
		&Method   & Success & Precision &FPS\\
		\hline
		&Coordinate Space\cite{qi2019deep}    & 70.3 & 81.9 &53.4\\
		&BEV Space  (w/o EEM) & 73.1 &\textbf{84.5} &50.1\\
		&BEV Space (w/ EEM)    & \textbf{73.4} &84.4 &\textbf{52.6}\\
		\hline
	\end{tabular}
		
	}

	\label{tab:8}%
\end{table}%

\begin{table}[t]
	\centering
	\large
	\caption{Integration with Siamese-based network.}
	
	\resizebox{\linewidth}{!}{
		\begin{tabular}{cccccc}
			\toprule
			Method & Car   & Pedestrian & Van   & Cyclist & Mean \\
			\hline
		STNet  & 72.1/84.0 & 49.9/77.2 & 58.0/70.6 & 73.5/93.7 & 61.3/80.1 \bigstrut[t]\\
		STNet* & 81.1/87.5 & 71.2/84.2 & 68.7/78.4 & 80.8/93.9 & 75.7/85.4 \\
		Improvement & \textcolor{green!70}{$\uparrow$9.0}/\textcolor{green!70}{$\uparrow$3.5} & \textcolor{green!70}{$\uparrow$21.3}/\textcolor{green!70}{$\uparrow$7.0} & \textcolor{green!70}{$\uparrow$10.7}/\textcolor{green!70}{$\uparrow$7.8} & \textcolor{green!70}{$\uparrow$7.3}/\textcolor{green!70}{$\uparrow$0.2} & \textcolor{green!70}{$\uparrow$14.4}/\textcolor{green!70}{$\uparrow$5.3} \\
			\midrule
			PTTR  & 65.2/77.4 & 50.9/81.6 & 52.5/61.8 & 65.1/90.5 & 58.4/77.8 \bigstrut[t]\\
			PTTR* & 74.1/87.2 & 72.3/88.2 & 58.8/70.5 & 73.2/93.4 & 72.0/86.3 \\
			Improvement & \textcolor{green!70}{$\uparrow$8.9}/\textcolor{green!70}{$\uparrow$10.2} & \textcolor{green!70}{$\uparrow$21.4}/\textcolor{green!70}{$\uparrow$6.6} & \textcolor{green!70}{$\uparrow$6.3}/\textcolor{green!70}{$\uparrow$8.7} & \textcolor{green!70}{$\uparrow$8.1}/\textcolor{green!70}{$\uparrow$2.9} & \textcolor{green!70}{$\uparrow$13.6}/\textcolor{green!70}{$\uparrow$8.5} \\
			\bottomrule
			\multicolumn{4}{l}{\small *Integrated the pre-training strategy.}
		\end{tabular}%
	}
	\label{tab:addlabel}%
\end{table}%

\begin{table}[t]
	\centering
	\caption{Comparison with different feature fusing modules. }
	\vspace{2mm}
		\begin{threeparttable}
	\begin{tabular}{cccc}
		\hline
		Method & Success & Precision & FPS \bigstrut\\
		\hline
		P2B-xorr\cite{qi2020p2b} & 71.5 & 82.4 & 39.4 \bigstrut[t]\\
		MLVSNet-xorr\cite{shan2021ptt} &70.6 & 81.4 &41.7 \\
		V2B-xorr\cite{hui20213d} & 71.8& 82.7 & 29.2 \\
		STNet-xorr\cite{zhou2022pttr} & 72.0 & 83.2 & 35.7 \\
		Easytrack++$^\dag$  & 72.4 & 83.5 & 49.8 \\
		Easytrack++ & \textbf{73.4} &\textbf{84.4}& \textbf{52.6}\bigstrut[b]\\
		\hline
	\end{tabular}
	\tiny              
	$^\dag$ represents the adoption of a mask strategy.  
	\end{threeparttable}
	\label{tab:5}%
\end{table}%

\noindent\textbf{3D pre-training.} To further investigate the effectiveness of the pre-training strategy, we conducted a series of pre-training epochs ranging from 0-600 to explore the impact of different pre-training durations on tracking performance. The results for different epochs within the Car category on the KITTI dataset are presented in Fig. \ref{fig:loss_pre}(a). The optimal performance was observed at epoch=300, with success/precision metrics reaching 87.9/90.1. Furthermore, to assess the generalization of the pre-training strategy, we integrated it into the PTTR framework. The Tab \ref{tab:addlabel} illustrates a significant performance improvement when pre-training is incorporated into the PTTR network (\textcolor{green!70}{$\uparrow$13.6}/\textcolor{green!70}{$\uparrow$8.5}).
Fig. \ref{fig:loss_pre}(c) presents a visualization of the influence of pre-training on convergence speed. It is evident that when utilizing pre-training, convergence is achieved as early as the 15th epoch. This advancement of approximately 20 epochs, compared to standard training, leads to a substantial reduction in training duration.

\begin{figure}[t]
	\begin{center}
		\includegraphics[width=\linewidth]{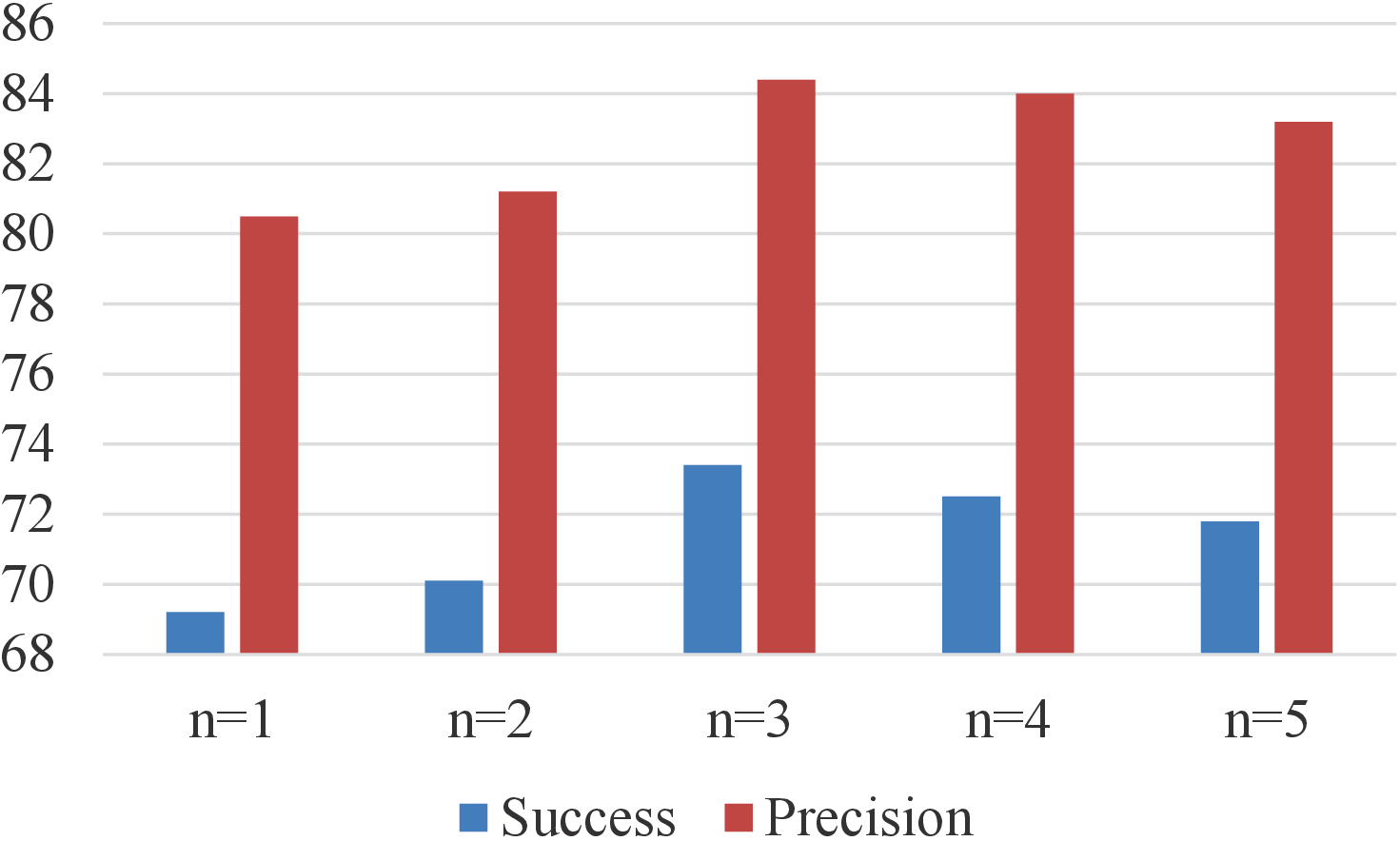}
	\end{center}
	\caption{Results of setting different $n$ values in the target-aware feature learning network.}
	\label{fig:11}
\end{figure}

\noindent\textbf{Layers of the transformer.} As shown in the Fig. \ref{fig:3}, We stack three layers of the self-perception transformer in the target-aware feature learning network in the EasyTrack++. The parameter $n$ of the self-perception transformer layers is a key parameter. If it is set too small, the ability of feature extraction and interaction of the network equilibrium point is insufficient.. If it is set too large, it will lead to overfitting and slow running speed. We report the results of setting different $n$ in the Car category in the KITTI dataset in Fig. \ref{fig:11}. We can find that as the parameter $n$ gradually increases, the tracking performance first increases and then decreases. The best performance is obtained when $n=3$. The Success/Precision metrics are 73.4/84.4.

\noindent\textbf{Joint or split point feature learning.} The core design of our tracker is a target-aware point cloud feature learning network. It effectively embeds the target's information into the search area and greatly accelerates the running speed. To verify the superiority of our design, we also design a typical Siamese tracker with the transform backbone in Fig. \ref{fig:4}(a). Different from EasyTrack++, we do not concatenate the template and search area before feature learning. Different feature fusing modules are equipped for a full comparison. As shown in Tab~\ref{tab:5}, we select four kinds of feature fusing networks including P2B-xorr\cite{qi2020p2b}, MLVSNet-xorr\cite{wang2021mlvsnet}, V2B-xorr\cite{hui20213d}, and STNet-xorr\cite{hui20223d}. We can find that EasyTrack++ achieves the best accuracy and fastest running speed. This demonstrates that the proposed target-aware feature learning network can improve the accuracy and running speed at the same time.

\noindent\textbf{Localization in the BEV or 3D feature space.} As discussed in Sec.~\ref{sec:3.3}, we regress and classify in a relatively dense BEV feature map instead of the 3D feature space. When facing sparse point clouds scenes, it is hard to generate high-quality proposals in the 3D space, so these methods may not be able to track the object effectively. To validate the superiority of our design, we preserve the feature learning network and replace the target localization network with the widely adopted VoteNet\cite{qi2019deep}-based strategy. However, due to the sparsity of point clouds, the voting stage suffers from heavy outliers and generates unreliable proposals. The results in the Car category in the KITTI dataset are shown in Tab ~\ref{tab:8}. We achieve better performance when regressing and classifying in the BEV feature space, 3.1/2.5 higher in the Success/Precision metrics. It further demonstrates the effectiveness of localization in the BEV feature space.

\noindent\textbf{Ways to generate center points.} In the proposed EasyTrack++, we design a center points interaction strategy to make secondary interaction to provide more detailed target information. The quality of center points has a significant impact on the tracking performance. In fact, we consider two ways to define the center points including ball query and K-nearest-neighbor (KNN) algorithms. These are two commonly used algorithms to capture local information in the point clouds scene. Meanwhile, the number of center points is also a key parameter. We conduct extensive ablation experiments in the Car category in the KITTI dataset. The results are illustrated in Tab~\ref{tab:6}. We can find that ball query shows better performance compared to KNN algorithm with the same number of points. And as the number of points increases, the tracking performance first increases and then stabilizes. Finally, we select ball query to crop center points and the number of points is set to 128. since it achieves the best tracking accuracy.

\begin{table*}[htbp]
	\centering
	\caption{Results of different template generating schemes. We report the Success/Precision metrics in the Car category in the KITTI dataset.}
	\begin{tabular}{ccccccccc}
		\hline
		Scheme & P2B & BAT & PTT & V2B & PTTR & STNet & EasyTrack & EasyTrack++ \bigstrut\\
		\hline
		First GT & 46.7/59.7 & 51.8/65.5 & 62.9/76.5 & 67.8/79.3 & 55.0/65.6 & 70.8/82.4 & 70.9/82.6 & \textbf{71.3}/\textbf{82.9} \bigstrut[t]\\
		Previous result & 53.1/68.9 & 59.2/75.6 & 64.9/77.5 & \textbf{70.0}/\textbf{81.3} & 65.0/77.1 & 66.0/76.6 & 69.6/80.1 & 69.8/80.3 \\
		First GT \& Previous result & 56.2/72.8 & 60.5/77.7 & 67.8/81.8 & 70.5/81.3 & 65.2/77.4 & 72.1/84.0 & 72.8/84.0 & \textbf{73.4}/\textbf{84.4} \\
		All previous results & 51.4/66.8 & 55.8/71.4 & 59.8/74.5 & 69.8/81.2 & 63.1/74.8 & \textbf{73.3}/\textbf{85.4} & 70.1/81.4 & 70.5/81.8 \bigstrut[b]\\
		\hline
	\end{tabular}%
	\label{tab:7}%
\end{table*}%

\begin{table}[t]
	\centering
	\caption{Results of different ways to generate the center points to make secondary interaction in the proposed EasyTrack++.}
	\begin{tabular}{cccc}
		\hline
		Method & Number & Success & Precision \bigstrut\\
		\hline
		\multirow{4}[2]{*}{KNN} & 32 & 72.4 & 83.6 \bigstrut[t]\\
		& 64 & 72.9 & 83.9 \\
		& 128 & 73.2 & 84.2 \\
		& 256 & 73.2 & 84.1 \bigstrut[b]\\
		\hline
		\multirow{4}[2]{*}{Ball Query} & 32 & 72.6 & 83.7 \bigstrut[t]\\
		& 64 & 73.1 & 83.9 \\
		& 128 & \textbf{73.4} & \textbf{84.4} \\
		& 256 & 73.2 & 84.1 \bigstrut[b]\\
		\hline
	\end{tabular}%
	\label{tab:6}%
\end{table}%

\noindent\textbf{Different template generating schemes.} The quality of the template has a great impact on the performance of the tracker since the target is given in the template. Incomplete template provides ambiguous target information. To find the best scheme, we consider four schemes to generate the template including using the first ground truth, using the previous result, using all previous results, and merging the first ground truth and previous result. The results of different schemes compared to other trackers in the Car category are shown in Tab~\ref{tab:7}. EasyTrack outperforms other trackers in most cases. And EasyTrack++ further improves the tracking performance in all schemes. However, we can find that when only using the previous result to generate template, V2B~\cite{hui20213d} achieves the best performance. It is because V2B uses an auxiliary shape completion network to capture more shape information. This makes the network structure more complex. When using all previous results, STNet\cite{hui20223d} achieves the best performance but at a very low speed. Considering the balance between speed and accuracy, all the trackers use the first ground truth and previous result to generate the template and our methods achieve the best performance in this scheme.

\section{Conclusions}
In this paper, we propose a new baseline for 3D single object tracking in point clouds abbreviated as EasyTrack, which can simplify the two-stream multi-stage 3D Siamese trackers, and unify the process of 3D feature extraction and target information integration, based on the masked point cloud tracking feature pre-trained learning. After that, we design a localization network in the BEV feature space for accurate regression and classification. In addition, we further design an enhanced version named EasyTrack++, which develops a center points interaction (CPI) strategy to reduce the ambiguous target description caused by the background information in the noisy point clouds. Extensive experiments in the KITTI, nuScenes and Waymo datasets demonstrate that EasyTrack achieves state-of-the-art performance and runs at a real-time speed. Due to the rich texture information provided by RGB data for incomplete point clouds, future plans include complementing point clouds with 2D images and proposing a multi-modal one-stream 3D tracker.


{\small
\bibliographystyle{IEEEtran}
\bibliography{MultiMedia}

}
\begin{IEEEbiography}
	[{\includegraphics[width=1in,height=1.25in,clip,keepaspectratio]{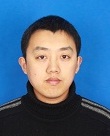}}]{Baojie Fan}
	is an professor in the Department of Automation college at Nanjing University of Posts and Telecommunications. He received the Ph.D. degree in pattern recognition and intelligent system from State Key Laboratory of Robotics, Shenyang Institute Automation, Chinese Academy of Sciences. His major research interest includes robot vision system, 2D/3D object tracking, and pattern recognition. He has published more than 20 top conferences and journals, such as IEEE TIP, TMM, TCSVT, PR, RAL, ICRA, ECCV, etc.
\end{IEEEbiography}

\begin{IEEEbiography}
	[{\includegraphics[width=1in,height=1.25in,clip,keepaspectratio]{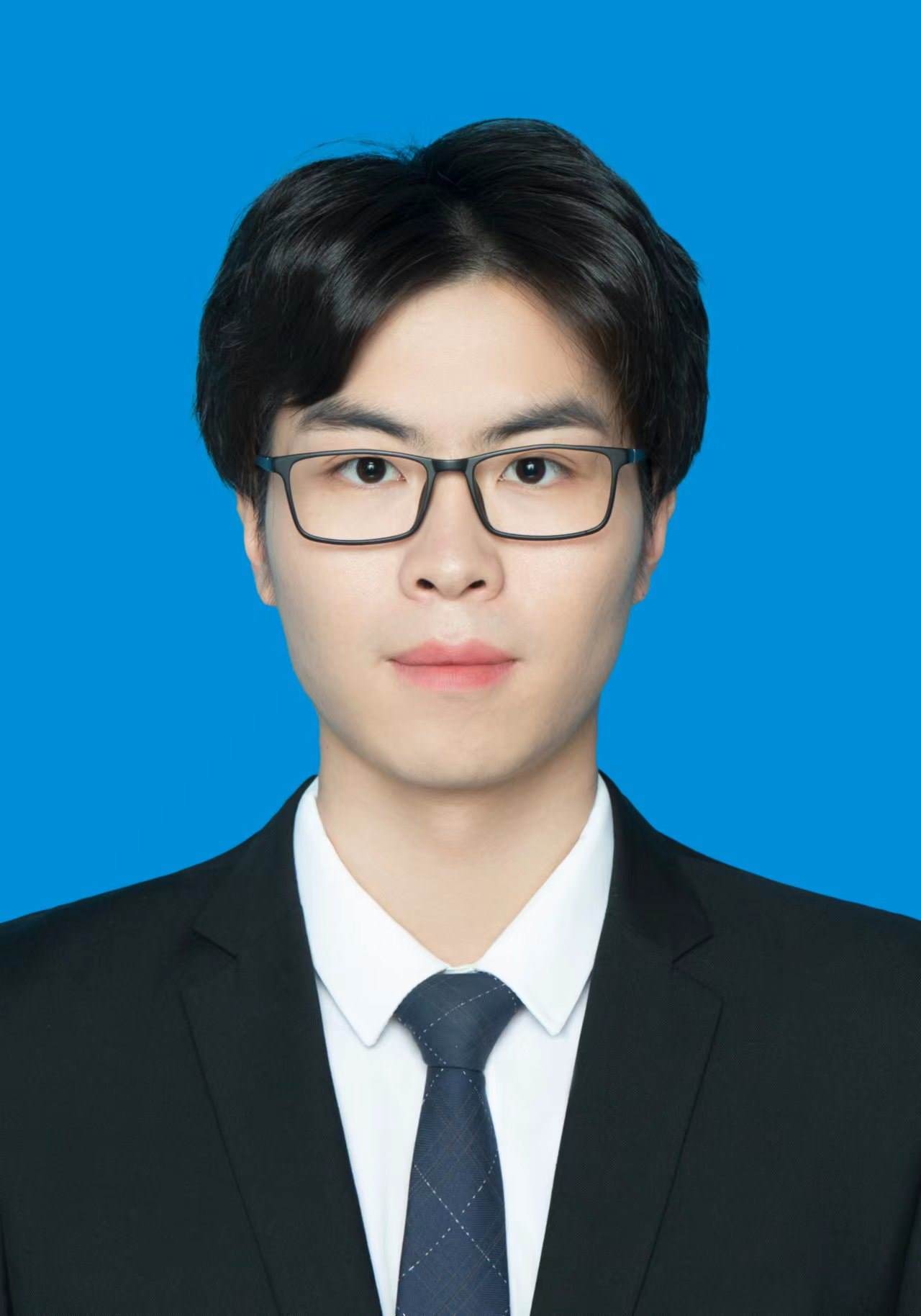}}]{Wuyang Zhou}
	is currently pursuing the Ph.D. degree with the Department of Automation college at Nanjing University of Posts and Telecommunications. His research focuses on point clouds analysis, 3D object tracking.
\end{IEEEbiography}

\begin{IEEEbiography}
	[{\includegraphics[width=1in,height=1.25in,clip,keepaspectratio]{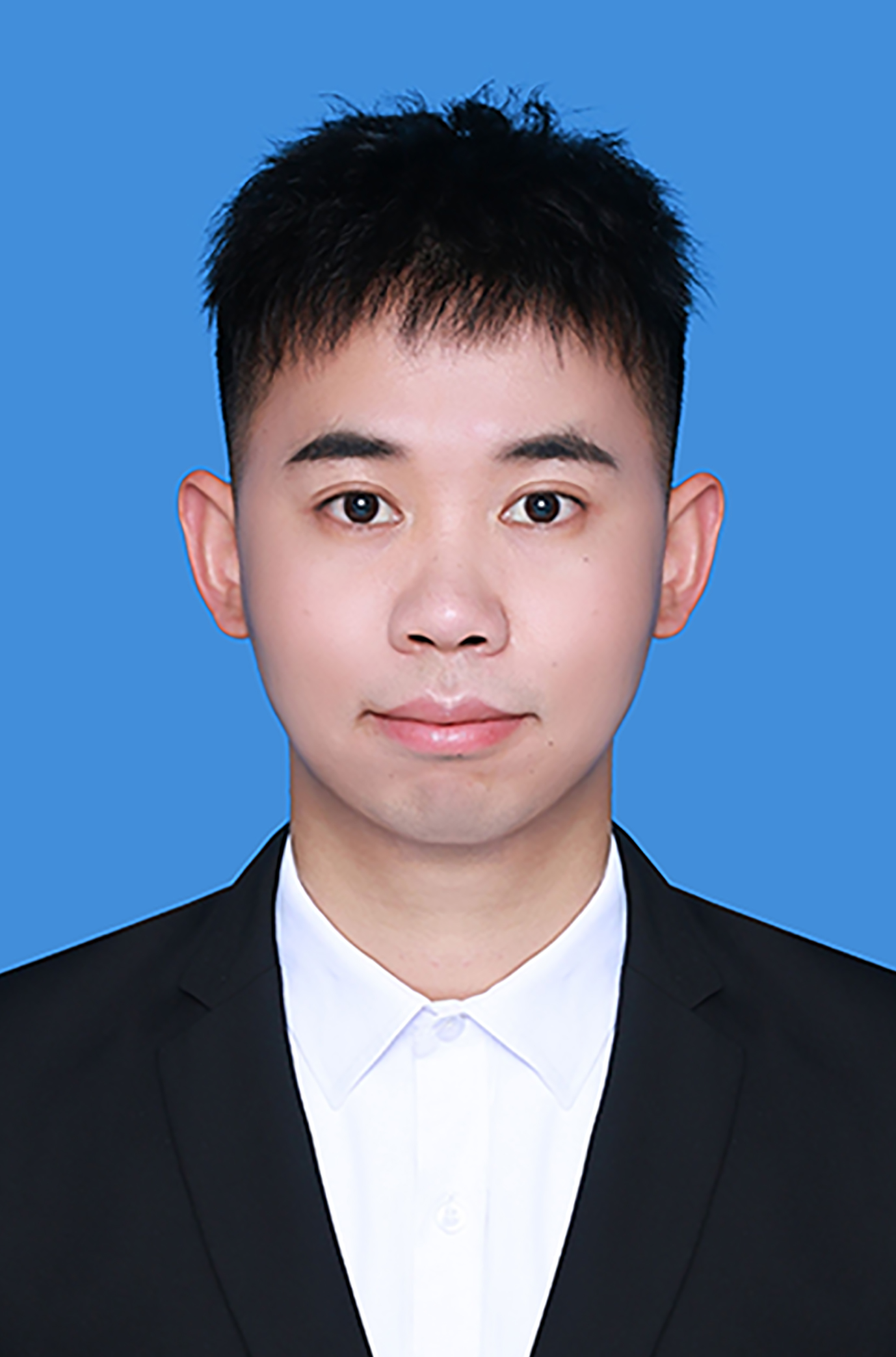}}]{Kai Wang}
	is currently pursuing the Ph.D. degree with the Department of Automation college at Nanjing University of Posts and Telecommunications. His research focuses on 3D object detection and tracking, multi-model object tracking. he has published multiple top conferences or journal papers, such as ICRA, RAL, etc.
\end{IEEEbiography}

\begin{IEEEbiography}
	[{\includegraphics[width=1in,height=1.25in,clip,keepaspectratio]{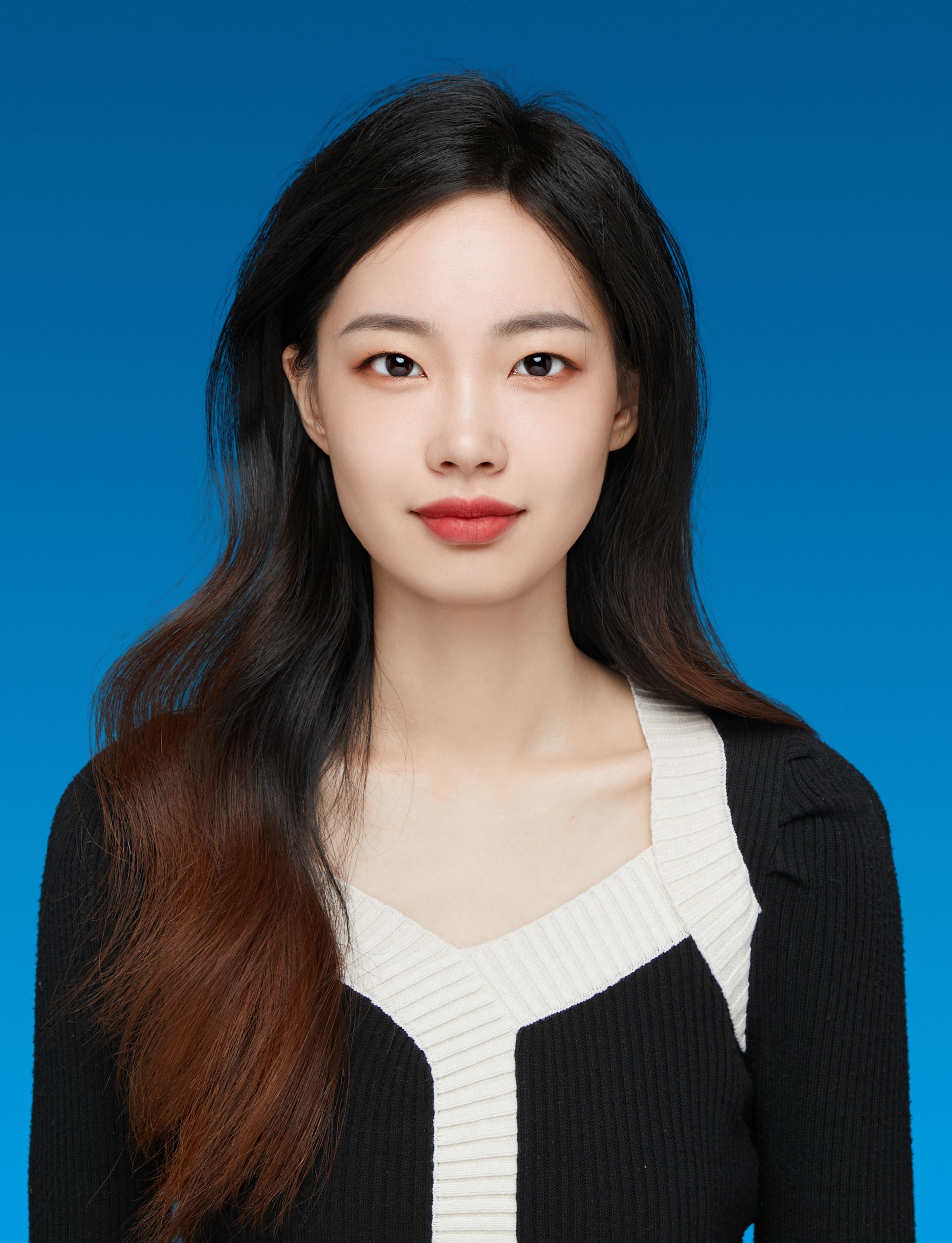}}]{Shijun Zhou}
	is currently pursuing the Ph.D. degree with the State Key Laboratory of Robotics, Shenyang Institute of Automation, University of Chinese Academy of Sciences. Her research focuses on computer vision methods in scattering media.
\end{IEEEbiography}

\begin{IEEEbiography}
	[{\includegraphics[width=1in,height=1.25in,clip,keepaspectratio]{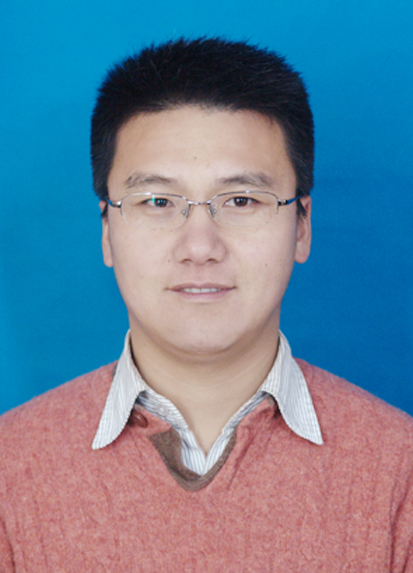}}]{Fengyu Xu} received the M.S. degree from Hefei University of Technology, Hefei, China, in 2005 and the Ph.D. degree from Southeast University, Nanjing, China, in 2009.
	From May 2016 to April 2017, he was a Visiting Scientist with the Department of Mechanical Engi neering, Michigan State University, East Lansing, MI, USA. He is currently a Full Professor, and the Associate Dean of the College of Automation and College of Artificial Intelligence, Nanjing Univer-sity of Posts and Telecommunications. His current research interests include inspection robot, soft robotics, machine vision and intelligent manufacturing.
	
\end{IEEEbiography}

\begin{IEEEbiography}
	[{\includegraphics[width=1in,height=1.25in,clip,keepaspectratio]{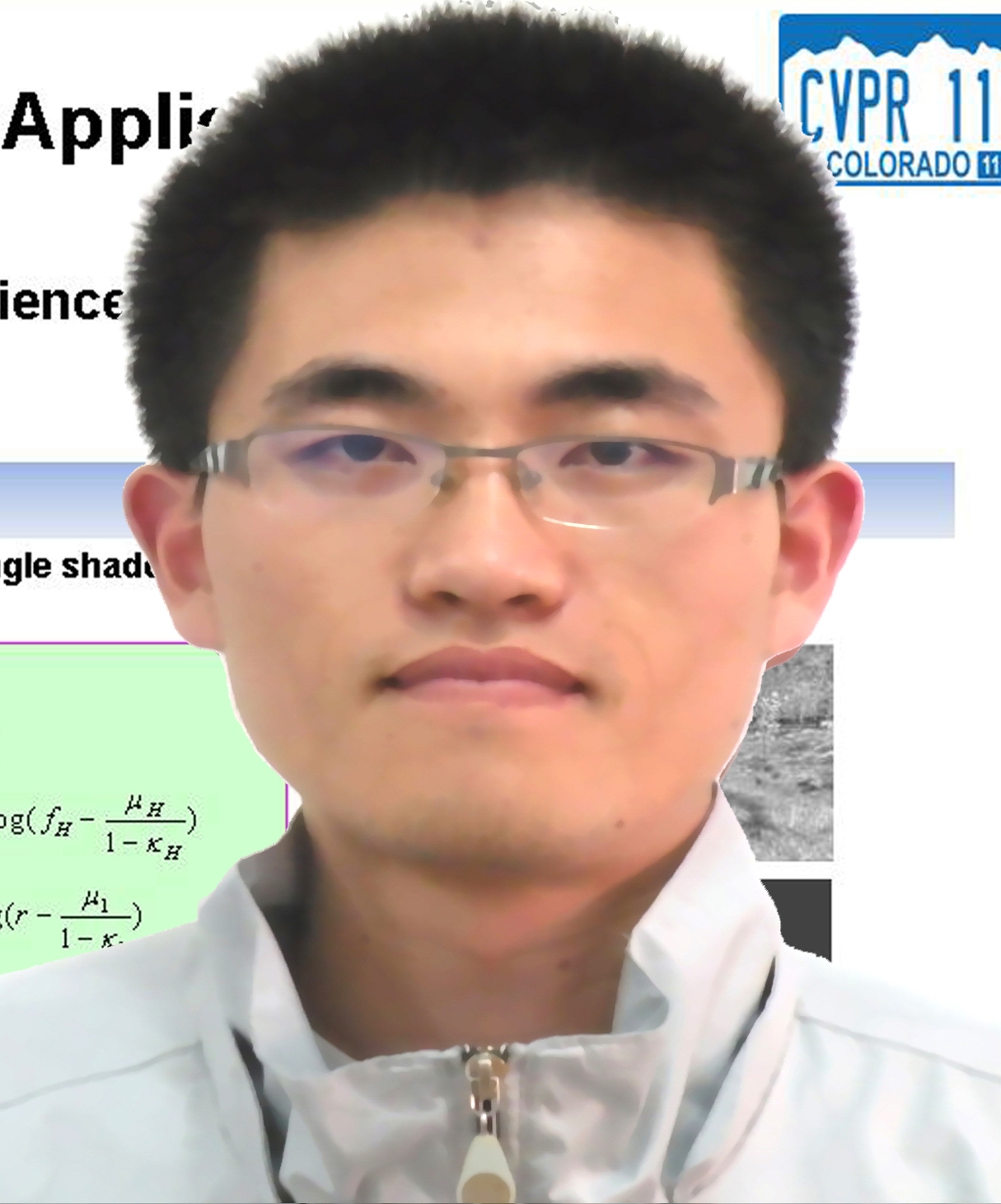}}]{Jiandong Tian}
	(Senior Member, IEEE) received the B.S. degree from the Department of Automation, Heilongjiang University, China, in 2005, and the Ph.D. degree from the Shenyang Institute of Automation, Chinese Academy of Sciences, in 2011. He is currently a Professor with the Shenyang Institute of Automation, Chinese Academy of Sciences. He has published more than 50 top journal and conference papers in the field of robot vision, image processing, and pattern recognition. He has also served as reviewer for several top
	journals and conferences such as T-PAMI, TIP, JMLR, TKDE, IJCV, TRO, CVPR, ICCV, ICRA, RSS and ECCV.
\end{IEEEbiography}

\end{document}